\documentclass[times,twocolumn,final]{elsarticle}

\usepackage{cag}
\usepackage{framed,multirow}
\usepackage{graphicx}
\usepackage{amssymb}
\usepackage{latexsym}
\usepackage{amsmath}
\usepackage{booktabs}
\usepackage{float}
\usepackage{url}

\usepackage{xcolor}
\definecolor{newcolor}{rgb}{.8,.349,.1}

\usepackage{hyperref}
\usepackage{subfigure}
\usepackage{amsmath}

\usepackage[switch,pagewise]{lineno} 

\journal{Computers \& Graphics}

\begin{document}

\verso{Preprint Submitted for review}

\begin{frontmatter}

\title{Terrain Point Cloud Inpainting via Signal Decomposition}%

\author{Yizhou Xie}
\author{Xiangning Xie}
\author{Yuran Wang}
\author{Yanci Zhang\corref{cor1}}
\author{Zejun Lv}

\cortext[cor1]{Corresponding author.}
\emailauthor{yczhang@scu.edu.cn}{Yanci Zhang}
\address{The College of Computer Science, Sichuan University, Chengdu 610065, China.}

    


\received{\today}

\begin{abstract}
The rapid development of 3D acquisition technology has made it possible to obtain point clouds of real-world terrains. However, due to limitations in sensor acquisition technology or specific requirements, point clouds often contain defects such as holes with missing data. Inpainting algorithms are widely used to patch these holes. However, existing traditional inpainting algorithms rely on precise hole boundaries, which limits their ability to handle cases where the boundaries are not well-defined. On the other hand, learning-based completion methods often prioritize reconstructing the entire point cloud instead of solely focusing on hole filling. Based on the fact that real-world terrain exhibits both global smoothness and rich local detail, we propose a novel representation for terrain point clouds. This representation can help to repair the holes without clear boundaries. Specifically, it decomposes terrains into low-frequency and high-frequency components, which are represented by B-spline surfaces and relative height maps respectively. In this way, the terrain point cloud inpainting problem is transformed into a B-spline surface fitting and 2D image inpainting problem. By solving the two problems, the highly complex and irregular holes on the terrain point clouds can be well-filled, which not only satisfies the global terrain undulation but also exhibits rich geometric details. The experimental results also demonstrate the effectiveness of our method.
\end{abstract}


\end{frontmatter}

\section{Introduction}
\label{sec::introduction}

\begin{figure}[h]
    \centering
    \subfigure[Raw data]{
    \includegraphics[width=0.48\linewidth]{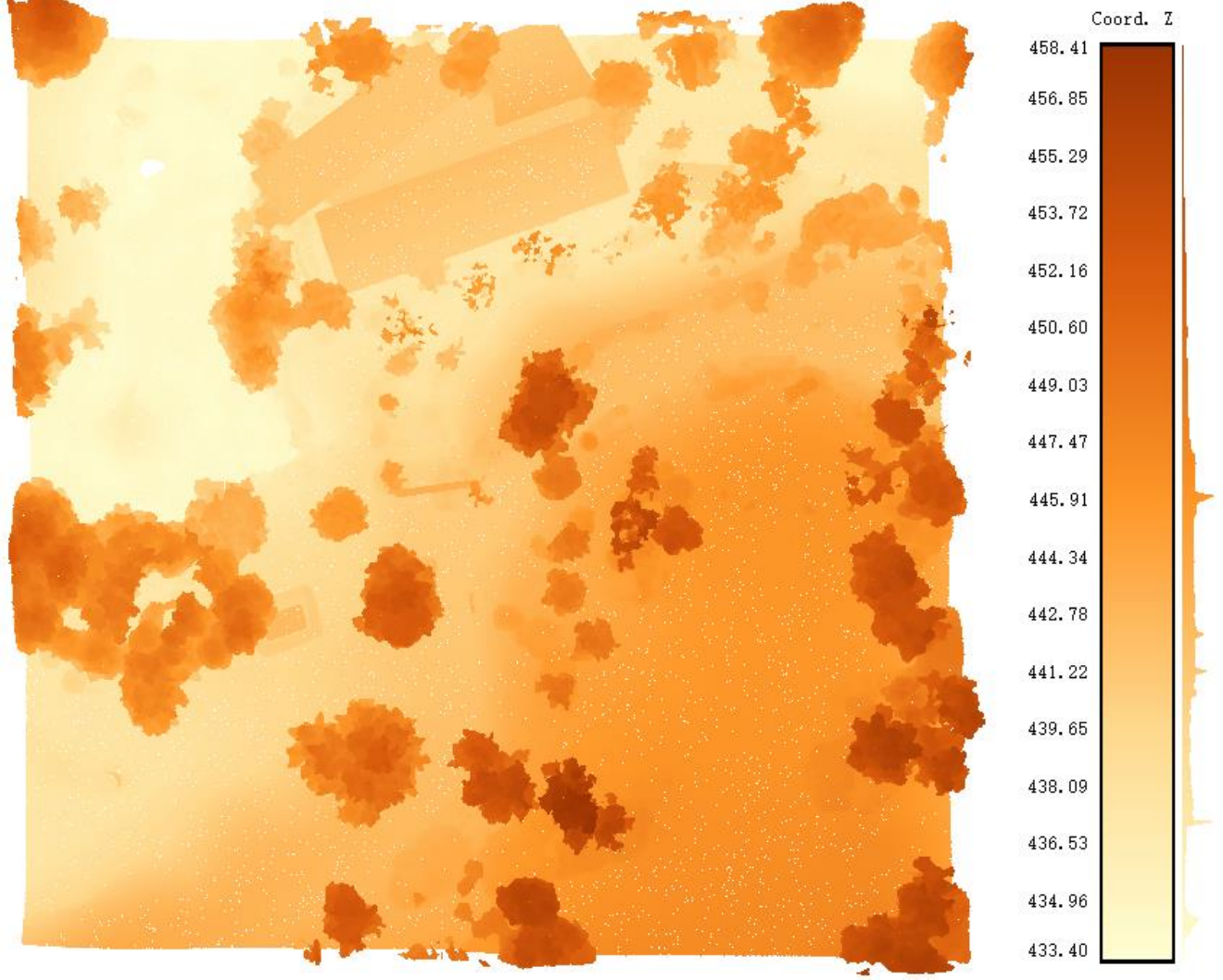}}
    \label{Fig::ShowHole_Hill_Raw}
    \subfigure[Holes with irregular shapes]{
    \includegraphics[width=0.48\linewidth]{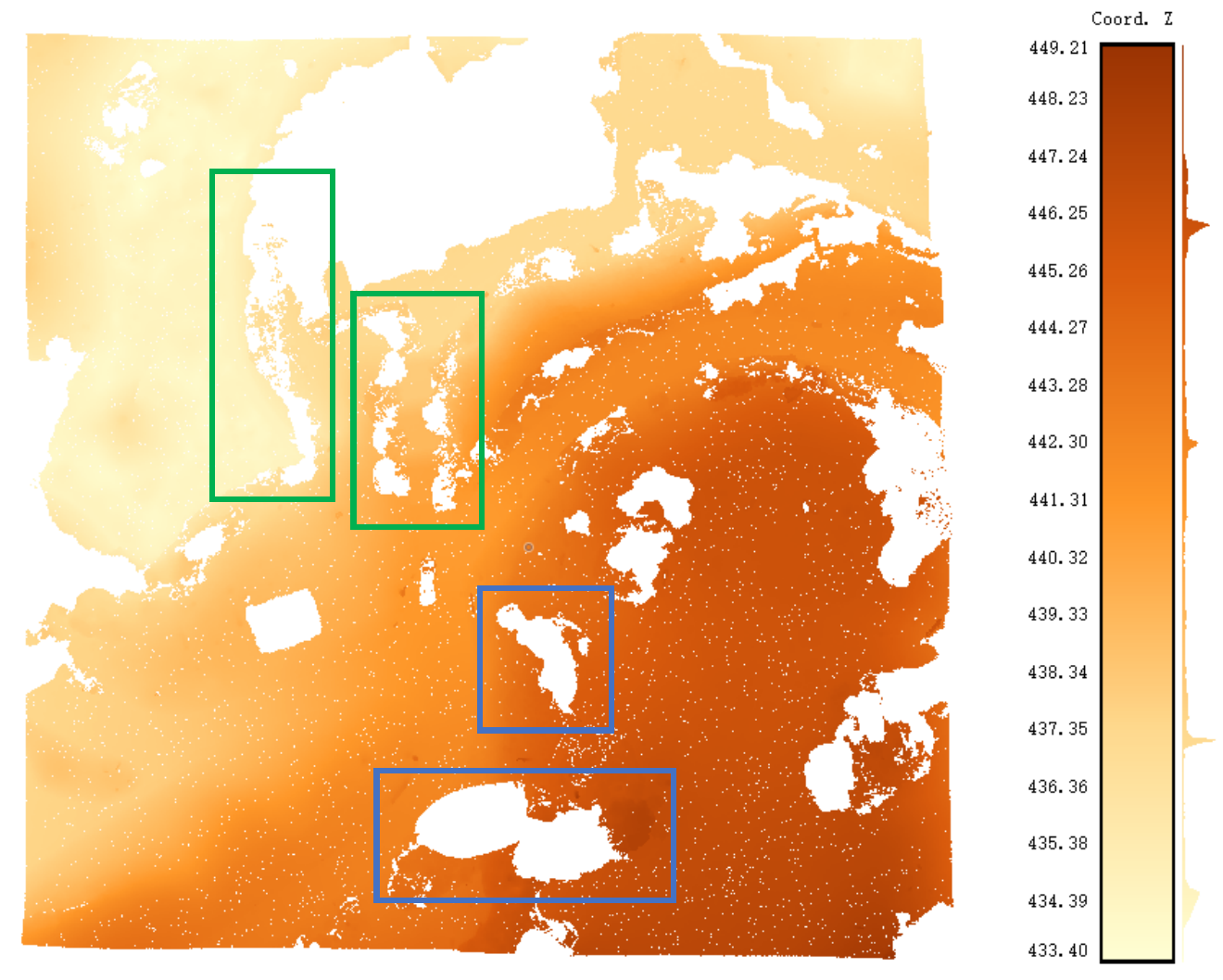}}
    \label{Fig::ShowHole_Hill_Seg}
    \caption{In outdoor scenes, the combination of trees and low-rise buildings often presents significant challenges for terrain reconstruction using a 3D acquisition technique called oblique photography. Upon filtering out these extraneous objects, complex-shaped or no-well-defined holes remain. (\textbf{a}) An overhead perspective of the terrain point cloud. (\textbf{b}) After applying filters, both vegetation and low-rise buildings are effectively eliminated, thereby exposing holes within the terrain. Irregular boundary holes are marked with blue boxes, while holes lacking well-defined boundaries are marked with green boxes.} \label{Fig::ShowHole_Hill}
\end{figure}

With the rapid advancements of modern sensor technology, it has become feasible to acquire point clouds of real-world terrains at an acceptable cost. However, holes may appear on the terrain due to the removal of undesirable objects or occlusion. As shown in Figure~\ref{Fig::ShowHole_Hill}, these holes may have irregular shapes, even without well-defined boundaries. \par

Currently, both traditional hole-filling algorithms and deep learning-based completion algorithms are dedicated to generating new point clouds from local point clouds. Unfortunately, existing traditional point cloud inpainting algorithms are not capable of handling these holes. Specifically, they often require a precise description of hole boundaries~\cite{ju20193d, fu2021dynamic}, and can only effectively operate on simple hole boundaries. However, the boundaries become so intricate that accurate determination in math becomes difficult for terrain point clouds, even unfeasible to define the boundaries. As a result, these algorithms cannot be effectively applied in this situation. \par

Moreover, while deep learning-based algorithms for point cloud completion demonstrate efficient performance on public release datasets such as ShapeNet-Part~\cite{chang2015shapenet}, PCN~\cite{yuan2018pcn} and Completion3D~\cite{tchapmi2019topnet}, they have challenges in handling terrain point clouds. Firstly, these methods are limited by the size of the neural network, making it difficult to generate a large number of points that match the scale of terrain point clouds. Secondly, these methods generate interpolated points through encoding and decoding the partial point cloud as a latent representation~\cite{yuan2018pcn}. Consequently, there is a tendency to reconstruct the entire point cloud instead of solely filling in the missing parts, resulting in a general reconstruction that lacks specific details~\cite{tabib2023defi}.

In this paper, we propose a novel method to repair complex-shaped holes on terrain point clouds. It is built on a new representation of terrain point clouds based on the fact that terrains can be treated as a combination of low- and high-frequency components, which represent the terrain undulation and geometric details respectively. With respect to the low-frequency components, we use B-spline surfaces to fit them. In addition, we build a relative height map based on the B-spline surfaces to represent the high-frequency components. Based on this new representation, the terrain point clouds inpainting problem is converted to a B-spline surface fitting problem and an image inpainting problem. Obviously, it is easier to fix these two problems than to patch complex-shaped point cloud holes directly. In addition, this method preserves terrain undulation and geometric details well by repairing both the low- and high-frequency information.

In summary, the main contributions of this paper are as shown below: 
\begin{itemize}
    \item We propose a new representation for terrain point clouds. It decomposes the terrain point clouds into low- and high-frequency components, which represent the terrain undulation and geometric details respectively.
    \item We design an automated hole location method. Specifically, we give a unique definition for holes and directly locate the position of the holes. In this way, the detection of complex hole boundary points is not required. 
    \item We fit the terrain point cloud to a B-spline surface and treat it as the low-frequency component. The use of the B-spline surface can help fully represent the local structural features of the 3D point cloud in its parameter space and avoid geometric loss in the format transformation.
\end{itemize}

The remainder of the paper is organized as follows. Section~\ref{sec::related_work} describes the related work. An overview of our method is presented in Section~\ref{sec::method_overview}. Section~\ref{sec::low-frequency_component_inpainting} describes the methods for constructing and inpainting low-frequency component. Section~\ref{sec::high-frequency_component_inpainting} presents our method of inpainting high-frequency components. Section~\ref{sec::reconstruction_from_decomposed_signals} introduces the process of reconstructing point cloud signals. Section~\ref{sec::handling_large-scale_point_clouds} provides a more detailed description when handling large-scale point clouds. The experimental results are shown in Section~\ref{sec::experiments}. The Discussion is provided in Section~\ref{sec::discussion}. Lastly, conclusions are given in Section~\ref{sec::conclusion}. \par

\section{Related Work}
\label{sec::related_work}

Currently, both deep learning-based methods and traditional methods have reached a mature stage. Deep learning-based point cloud completion methods have made rapid advancements, encompassing point-based~\cite{ding2021rfnet}, convolution-based~\cite{wang2020softpoolnet}, graph-based~\cite{zhang2021pc}, transformer-based~\cite{xiang2021snowflakenet}, and generative model-based~\cite{cheng2021dense}. These techniques have demonstrated impressive results on existing datasets in the CAD field, such as ShapeNet~\cite{chang2015shapenet}, ModelNet~\cite{wu20153d}, and Completion3D~\cite{tchapmi2019topnet}. However, their robustness is limited when confronted with real-world scenarios that lack an abundance of actual data. Furthermore, the constraint of GPU memory hinders direct processing of large-sized point clouds using deep learning methods. Voxelization preprocessing, although utilized, can lead to the loss of essential information within the point cloud. Despite the existence of feature extraction networks like PointNet~\cite{qi2017pointnet}, there is a need to further emphasize feature learning~\cite{fei2022comprehensive}.

Based on the data format, existing traditional point cloud inpainting methods can be broadly classified into two categories: mesh-based inpainting methods, and point-based inpainting methods. Although mesh format applied in mesh-based inpainting methods offers a natural advantage in automatically detecting hole boundaries due to its inherent topology, the geometric loss incurred during the inpainting and format transformation process is unacceptable. \par

\textbf{Mesh-based inpainting methods.} Generally, these methods firstly grid point clouds, and then detect hole edges based on mesh topology. Lastly, they use the geometric structure and topological characteristics of the neighborhood to interpolate the hole area~\cite{liepa2003filling, sahay2015geometric}. For example, Wei \textit{et al.}~\cite{wei2010integrated} applied a minimal triangulation to holes and iteratively subdivided the patching meshes in order to assort with the density of surrounding meshes. However, mesh interpolation can become trapped in planar hole-filling content~\cite{fu2018point}, resulting in a lack of detail. In addition to these grid-based methods for filling holes in general models, there are also unique methods for oblique photographic data. For example, Wang \textit{et al.}~\cite{wang2021parallel} divided the point cloud space into cells and individually repaired each cell to accelerate the hole-filling process. However, this method encounters precision issues when the reconstructed mesh is rough and performs poorly when dealing with large holes. \par

\begin{figure*}
        \centering
        \includegraphics[width=0.8\linewidth]{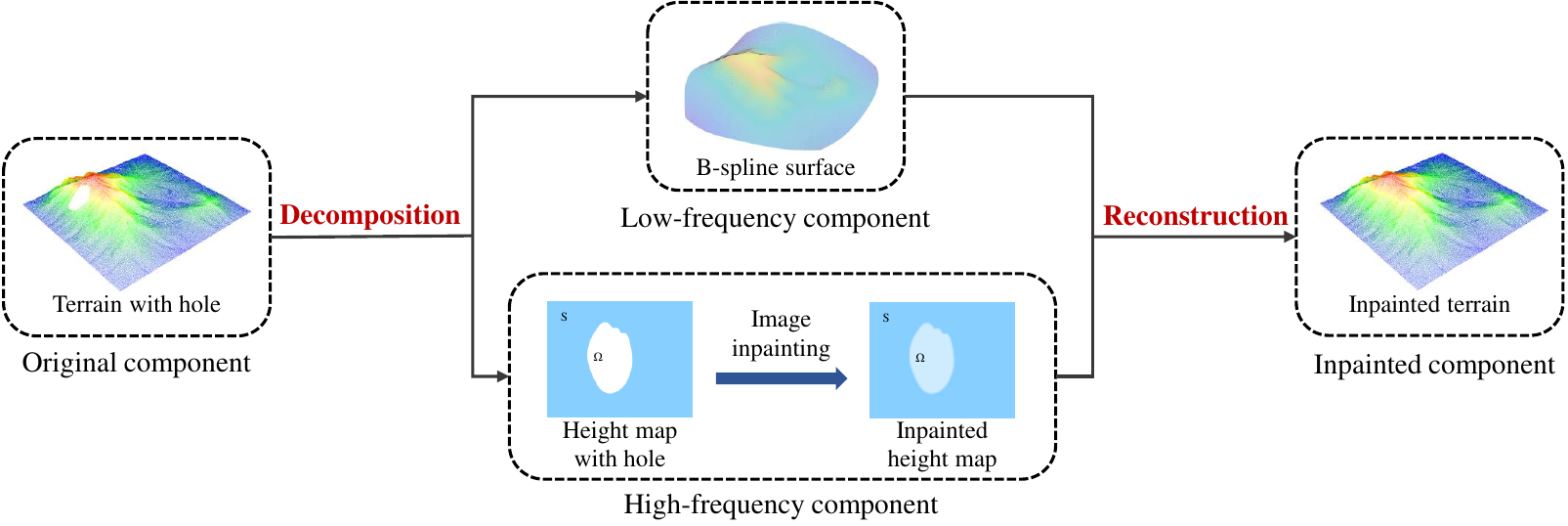}
    \caption{Core idea of our method.}
    \label{Fig::BasicIdea}
\end{figure*}

\textbf{Point-based inpainting methods.} These point cloud inpainting methods work directly on point clouds, avoiding the costly process of format transformation. One common solution for inpainting is the patch-based matching scheme. For example, Huang \textit{et al.}~\cite{huang2023plgp} detected holes directly in point clouds and then iteratively searched for context with local similarity, propagating it appropriately along the local geometric structure of occlusion. Dinesh \textit{et al.}~\cite{dinesh2017exemplar} filled holes in an iterative way. Specifically, they gave boundary points priority determining the order in which they were processed, and used templates near the hole boundary to find the best matching regions. Additionally, Shi \textit{et al.}~\cite{shi2022point} considered the normal features of point clouds for patch matching and replacing. Some works attempt to simplify the problem of detecting holes and inpainting by adjusting the representation of unorganized point clouds. For example, Doria \textit{et al.}~\cite{doria2012filling} projected points onto a depth map from a certain perspective and then used an image-based method to complete the inpainting and reconstruct 3D point clouds from the inpainted image. Hu \textit{et al.}~\cite{fu2018point, hu2019local} organized point clouds with a graph and proposed a method that exploits both the local smoothness and non-local self-similarity, which also relies on the depth map to find holes. However, while these methods introduce new representations and work well in many cases, they can lead to geometric losses in constructing depth maps when facing complex scenarios. Altantsetseg \textit{et al.}~\cite{altantsetseg2017complex} created contour curves inside the boundary edges of the hole to add locally uniform points to the hole, but this method heavily relies on accurate hole boundaries. \par 


\section{Method Overview}
\label{sec::method_overview}

Our method proposes a new point cloud representation for terrain, based on the important fact that terrain possesses globally smooth with rich local geometric details~\cite{yu2021integrated}. Depending on this representation, we have the capability to process terrain point clouds of arbitrary shapes and the problem of filling holes in 3D point clouds can be primarily transformed into a 2D image inpainting problem. Moreover, there is no longer a need to extract hole boundary points as required by previous methods; instead, we can directly locate the hole regions. \par

\begin{figure} [h]
    \centering
    \subfigure[DEM Method]{
    \includegraphics[width=0.4\linewidth]{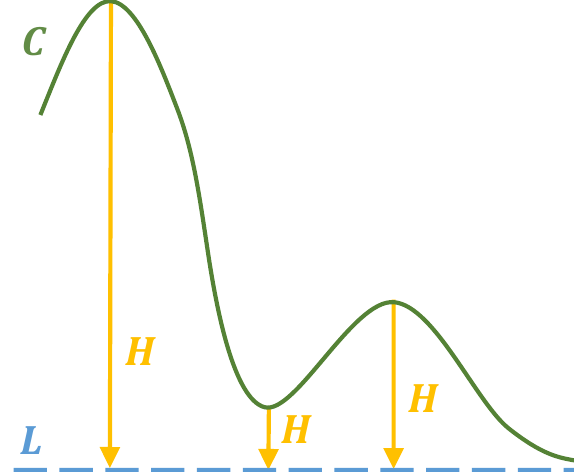}}
    \label{Fig::Comparison_DEM}
    \subfigure[Ours]{
    \includegraphics[width=0.4\linewidth]{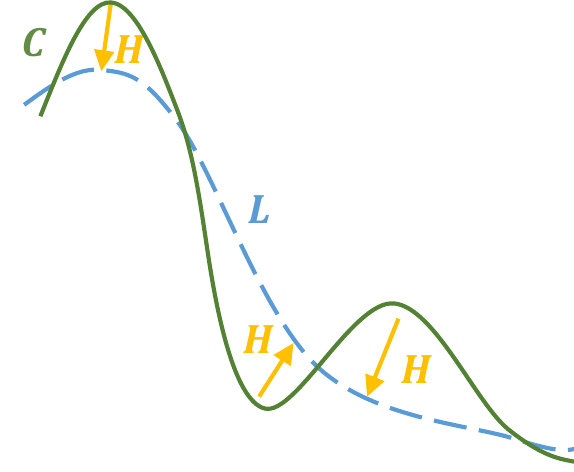}}
    \label{Fig::Comparison_Ours}
    \caption{A comparison is made between the DEM method and our method from a lateral perspective. In this comparison, the original signal is depicted in green, the low-frequency component $L$ is represented in blue, and the high-frequency component $H$ is depicted in yellow}. (\textbf{a}) illustrates the DEM method, where the $XY$ plane corresponds to the ground plane and represents $L$, while the height of the original signal above the ground represents $H$. (\textbf{b}) illustrates our algorithm, where the fitting B-spline surface is considered as $L$, while the distance between the original signal and the surface represents $H$.
    \label{Fig::Comparison_Overview}
\end{figure}

As shown in Figure~\ref{Fig::BasicIdea}, the core idea of our method is to decompose terrain point clouds $\mathcal{C}$ into a low-frequency component $L$ and a high-frequency component $H$, where $L$ describes the basic shape of the terrain and $H$ captures the details based on $L$. It is worth mentioning that DEM (Digital Elevation Model) in geographic information systems can also be applicable within our new representation. As illustrated in Figure~\ref{Fig::Comparison_Overview}, it could be treated as the simple form of the above decomposition, where the $XY$ plane is fixedly chosen as $L$ and the elevation is considered as $H$. However, $XY$ plane is not a good choice in every situation to represent the low-frequency information of terrain because it provides no information about the terrain undulations. \par

\begin{figure*}[h]
    \centering
    \includegraphics[width=0.9\linewidth]{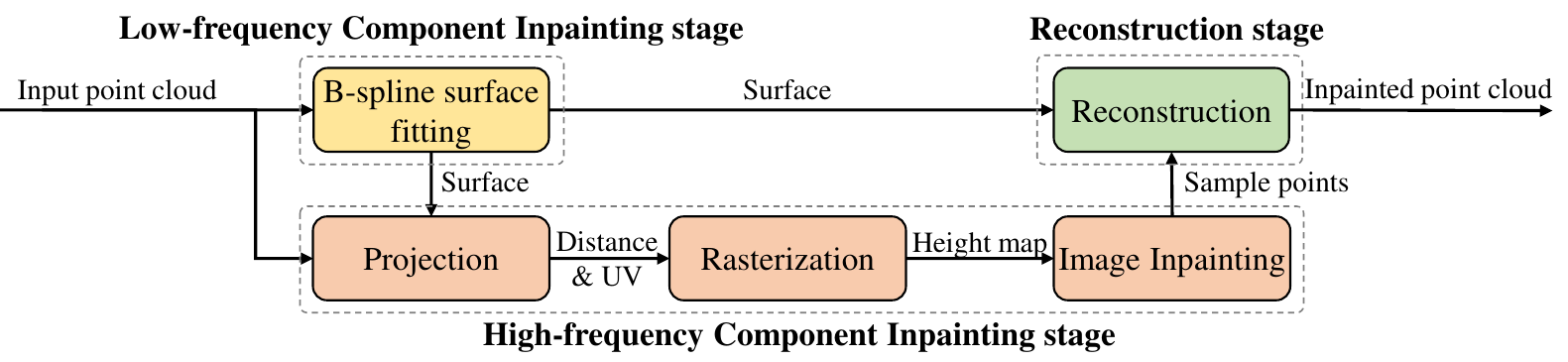}
    \caption{The pipeline of our method.}
    \label{Fig::Overview pipeline}
\end{figure*}

As shown in the Figure~\ref{Fig::Overview pipeline}, the proposed method consists of three stages and the detail description of them is shown below: \par

\begin{itemize}
    \item \textbf{Low-frequency Component Inpainting Stage:} In this stage, we utilize a B-spline surface as the low-frequency component to fit the 3D point cloud. Once the fitting process is complete, we consider the inpainting of the low-frequency component to be finished.
    \item \textbf{High-frequency Component Inpainting Stage:} We generate the high-frequency component by projecting each point to the B-spline surface, which involves finding the nearest point on the surface and calculating the projection distance. We record the projection information above to construct a high-frequency image (i.e., height map) after rasterizing. Then, we fill holes in the height map by solving the Poisson equation, guided by patch matching in the gradient domain.
    \item \textbf{Reconstruction Stage:} We compute offset distances away from the low-frequency B-spline surface by sampling in the high-frequency image, and then resolve 3D points by superimposing these two components to complete the reconstruction process. 
\end{itemize}

\section{Low-frequency Component Inpainting}
\label{sec::low-frequency_component_inpainting}

B-spline surface fitting is a well-established technique with a solid theoretical foundation and logical framework. It has been widely used in the field of reverse engineering~\cite{ma1998nurbs,brujic2011fast}. Generally, B-spline surface $\mathcal{S}(u,v)$ can be defined as a network of tensor product B-spline surface as shown in Equation~(\ref{equ::b-spline_surface}):

\begin{equation}
 \mathcal{S}(u,v) = \sum_{i=0}^m\sum_{j=0}^nN_{i,d_x}(u)N_{j,d_y}(v)B_{i,j}  \label{equ::b-spline_surface}
\end{equation}
where $B$ denotes a set of $m+1$ rows and $n+1$ columns control points, and $B_{i,j}\in\mathbb{R}^3$ represents the control point at $i$-row and $j$-column. $N_{i,d_x}(u)$ and $N_{j,d_y}(v)$ are B-spline basis functions of degree $d_x$ and $d_y$, defined over the knot sequences $u = (u_0, u_1, \cdots, u_{m+d_x})$ in the $u$-direction and $v = (v_0, v_1, \cdots, v_{n+d_y})$ in the $v$-direction respectively, with local coordinates $(u,v)\in[0,1]^2$. \par

The goal of B-spline surface fitting is to identify the control points $B$ and knot sequences $(u,v)$ that minimize the distance of $\mathcal{C}$ to $\mathcal{S}$, as shown in Equation~(\ref{equ::b-spline_surface_distance}):

\begin{equation}
D_{\mathcal{C}\rightarrow \mathcal{S}} = \sum_{i=1}^{N_C}{d(p_i, \mathcal{S})}^2  \label{equ::b-spline_surface_distance}
\end{equation}
where $N_\mathcal{C}$ is number of points in $\mathcal{C}$. $d(p_i, \mathcal{S})$ represents the distance of discrete point $p_i$ in $\mathcal{C}$ to B-spline surface $\mathcal{S}$ which can be calculated by Equation~(\ref{equ::point_to_b-spline_surface_distance}):

\begin{equation}
 d(p_i,\mathcal{S}) = \mathop{\min}_{(u_i,v_i)\in[0,1]^2} \parallel p_i-\mathcal{S}(u_i,v_i)\parallel^2  \label{equ::point_to_b-spline_surface_distance}
\end{equation}
where $(u_i,v_i)$ is the parametrization of the projection of $p_i$ onto $\mathcal{S}$.

To simplify the fitting problem, $(u,v)$ is obtained by uniform parameterization. Obviously, Equation~(\ref{equ::b-spline_surface_distance}) is a nested optimization problem, and can be naturally solved by an iterative optimization method. Specifically, each iteration is composed of two steps: 
 
\begin{itemize}
    \item \textbf{Fitting step.} For the given $(u_i,v_i)$ of projection point of ${p}_i$, the best $B$ are obtained by solving a linear least squares problem.
    \item \textbf{Correcting step.} For the given $B$, the best $(u_i,v_i)$ of projection point of $p_i$ are obtained by projecting $p_i$ onto the given $\mathcal{S}$.
\end{itemize}

It is worth noting that the B-spline surface fitting method mentioned above is based on iterative local optimization. There is no guarantee of convergence to a global optimum. The success of obtaining an optimal approximation depends on a good choice of the initial B-spline surface shape. However, if the target shape is relatively uncomplicated, a simple initial surface is generally enough to achieve stable convergence. Coincidentally, the natural terrain point clouds in our task are not particularly complex from an overall perspective. Therefore, we use a simple method of initializing $B$ by building an octree~\cite{wang2006fitting}, which balances both time cost and fitting quality. After initialization, only a few iterations are typically necessary to significantly improve the fitting accuracy. 

After getting $B$ by the above iteration optimization method, we can obtain the fitted B-spline surface $L$, i.e., the inpainted low-frequency component.

\section{High-frequency Component Inpainting}
\label{sec::high-frequency_component_inpainting}

\subsection{High-frequency Component Construction}
\label{sec::high-frequency_component_construction}
Once the low-frequency component $L$ has been constructed, the calculation of the high-frequency component $H$ follows naturally. To be more precise, $H$ corresponds to the signed distance of $\mathcal{C}$ projected onto $\mathcal{S}$. \par

\subsubsection{Projection Point Calculation}
\label{sec::projection_point_calculation}

If a point $q_i$ ($q_i = \mathcal{S}(u_i,v_i)$) on the surface minimizes the distance $d(p_i, q_i)$, we consider $q_i$ as the projection of $p_i$ onto the surface. In order to calculate $q_i$, Newton's method is employed to solve this problem. Specifically, the coordinates $(u^*_i,v^*_i)$ that minimize $d(p_i,\mathcal{S}(u_i,v_i))$ satisfy the condition $d^{\prime}(p_i,\mathcal{S}(u_i,v_i))=0$. To find the root of this equation, we can utilize Newton's method, which involves the iteration formula:

\begin{equation}
(u^{*,m+1},v^{*,m+1}) = (u^{*,m},v^{*,m}) - \frac{d^{\prime}(p_i,T_{u^{*,m},v^{*,m}})}{d^{\prime\prime}(p_i,T_{u^{*,m},v^{*,m}})}, m=0,1,2,\cdots \label{equ::newton's_method}
\end{equation}
Here, $m$ represents $m$-th iteration. This method exhibits quadratic convergence, indicating that Newton's method is generally faster than quadratic minimization and can converge to a solution within a few iterations~\cite{wang2002robust}. Initially, the method may have a slower start, but it progressively approaches the optimal value and rapidly converges to the final solution $d(p_i,q_i^{\prime})$, where $q_i^{\prime}$ represents the optimal projection point of $p_i$. \par

\subsubsection{Projection Point Initialization}
\label{sec::projection_point_initialization}

It is important to note that the choice of initial values in Newton's method significantly influences the efficiency of convergence. To address this, we propose a method for calculating the initial projection point $q_i^0$($q_i^0 =  \mathcal{S}(u_i^0,v_i^0)$) prior to executing iterations, thereby accelerating the subsequent steps. Firstly, we sample $\mathcal{S}$ with a uniform 2D sampling sequence $\{(u_0,v_0),(u_0,v_1),\dots,(u_\zeta,v_\eta)\}$, where $(u_k,v_l)=(\frac{k}{\zeta},\frac{l}{\eta})$, $a\in[0,\zeta]$, $b\in[0,\eta]$. These sample points are represented as $\mathcal{S}(u_k,v_l)$. Secondly, for each data point $p_i$, we calculate the distances $d(p_i,\mathcal{S}(u_k,v_l))=\parallel p_i-\mathcal{S}(u_k,v_l)\parallel$ from $p_i$ to $\mathcal{S}(u_k,v_l)$. Subsequently, we select the minimum $d(p_i,\mathcal{S}(u_k,v_l))$, denoted as $d(p_i,\mathcal{S}(u_k^{\prime},v_l^{\prime}))$ and consider $q_i^0 = \mathcal{S}(u_k^{\prime},v_l^{\prime})$ as the initial projection point of $p_i$. \par

\subsubsection{Sign Determination of Projection Distances}
\label{sec::sign_determination_of_projection_distances}

\begin{figure}[h]
    \centering
    \includegraphics[width=1.0\linewidth]{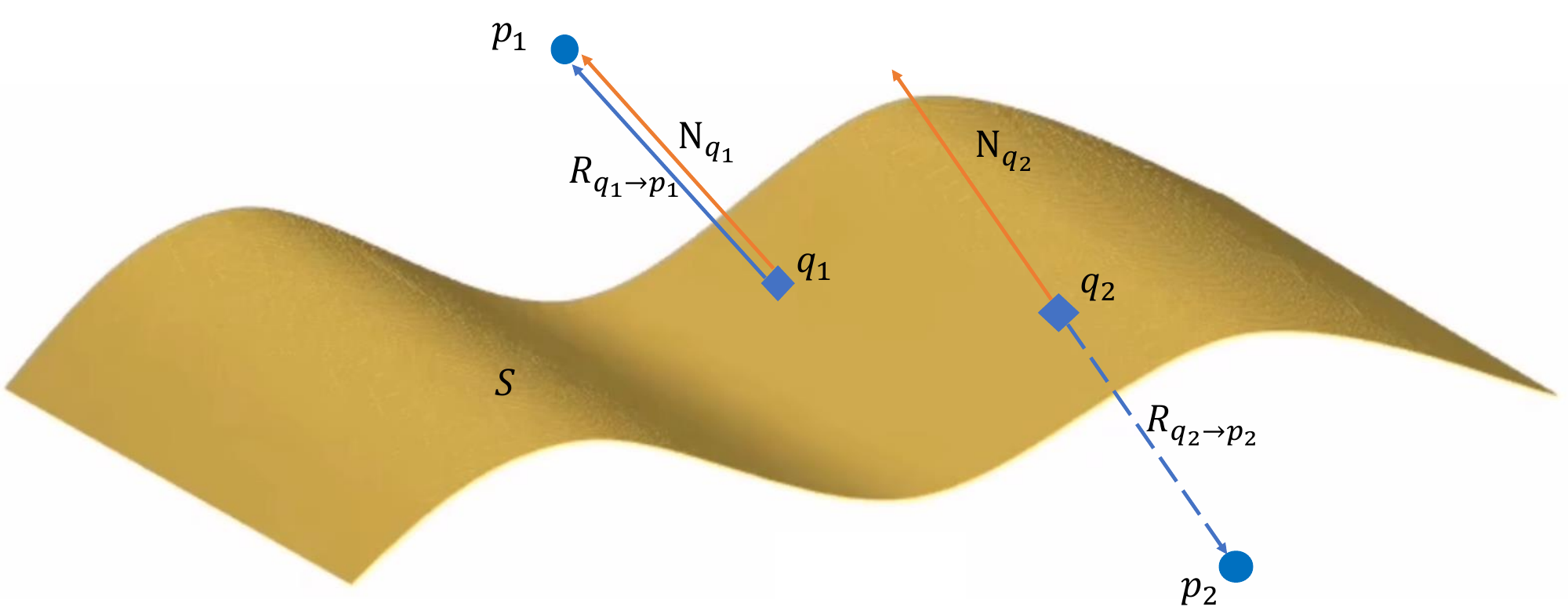}
    \caption{Illustration of point projection onto a surface. $q_1$ and $q_2$ represent the projection points of $p_1$ and $p_2$, respectively. $p_1$ and $p_2$ reside on opposite sides of $\mathcal{S}$ and exhibit different projection directions.}
    \label{Fig::point_projection}
\end{figure}

To ensure accurate generation of the point cloud in the subsequent reconstruction step, it is crucial to differentiate between the two potential directions, as depicted in Figure~\ref{Fig::point_projection}. In simpler terms, we utilize the normal vector $\vec{n}_{q_i}$ at $q_i$ and the vector from $q_i$ to $p_i$, denoted as $\vec{n}_{q_i} \cdot R_{q_i\rightarrow p_i}$ to determine this distinction. Theoretically, since $q_i$ is the closest point to $p_i$ on $\mathcal{S}$, the line that encompasses $\vec{n}_{q_i}$ must pass through $p_i$, resulting in two possibilities for the orientation of $\vec{n}_{q_i}$ and $R_{q_i\rightarrow p_i}$: they can either be parallel or anti-parallel. However, considering that in Section~\ref{sec::projection_point_calculation} we obtained $q_i$ through a finite number of iterations using Newton's method, it becomes necessary to account for the presence of an error $\epsilon$. In other words, the condition that holds true is:

\begin{equation}
1 - \lvert \vec{n}_{q_i} \cdot R_{q_i\rightarrow p_i} \rvert < \epsilon
\end{equation}
where $\vec{n}_{q_i}$ and $R_{q_i\rightarrow p_i}$ represent unit vectors. Additionally, we can assume that $q_i$ are highly reliable, indicated by $\lvert \vec{n}_{q_i} \cdot R_{q_i\rightarrow p_i} \rvert > 0$. \par

Therefore, the calculation of $\vec{n}_{q_i}$ is necessary as $\mathcal{S}$ itself does not include normal information. To achieve this, we gather all $q_i$ as a new point cloud and construct a \textit{K}-nearest neighbors (\textit{K}-NN) graph for them. Subsequently, we employ Principal Component Analysis (PCA)~\cite{martinez2001pca} to estimate their respective normals $\vec{n}_{q_i}$. Since the orientation of $\vec{n}_{q_i}$ can be ambiguous, a broadcasting method is applied as a post-processing step to ensure consistent orientation across all $\vec{n}_{q_i}$. Once this is done, we can proceed to calculate the sign of the projection distance denoted as $\omega$:

\begin{equation}
\omega_i = \frac{\vec{n}_{q_i} \cdot R_{q_i\rightarrow p_i}}{\lvert \vec{n}_{q_i} \cdot R_{q_i\rightarrow p_i} \rvert}
\label{equ::normal_direction}
\end{equation}
Furthermore, the calculation of the signed distance, specifically the high-frequency components $H$, can be obtained using the following procedure: 

\begin{equation}
H = D_{p_i} = \omega_i \times d(p_i,q_i) \label{equ::signed_distance}
\end{equation}

\subsection{Height Map Construction}
\label{sec::automatic_hole_location}

Considering the inconvenience of directly locating holes from projection information, which hinders the high-frequency component inpainting, we construct a height map to achieve this purpose. By projecting point cloud $\mathcal{C}$ onto $\mathcal{S}$ and generating projection points $q_i$ ($q_i = \mathcal{S}(u_i,v_i)$) for each corresponding point $p_i$ in $\mathcal{C}$, we establish a one-to-one correspondence between $(u_i,v_i)$ and $p_i$. This means that the projection information in the two-dimensional $UV$ space can reflect the local neighborhood relationships in the three-dimensional point cloud space. While directly locating holes or extracting boundaries in the continuous two-dimensional space is not convenient, our primary interest lies in determining the hole positions rather than identifying the points that belong to the hole boundaries. Therefore, we discrete the $UV$ space by rasterizing it into pixels at a self-adaptive resolution $(r,r)$, and examine the presence of $q_i$ within these pixels to determine their classifications and values. In simpler terms, pixels without any $q_i$ are considered holes, while pixels containing $q_i$ undergo further analysis to determine their values. \par

\subsubsection{Self-adaptive Resolution Estimation}
\label{sec::self-adaptive_resolution_estimation}

For convenience, we denote the corresponding coordinates of $q_i$ in the two-dimensional $UV$ space as $t_i$, where $t_i = (u_i,v_i)$. Based on empirical observations, we find that the local value of $r$ should be closely associated with the local density of $t_i$, denoted as $\rho$. Assuming that the global density of $t_i$ remains relatively constant, we propose a simple method to estimate it. Firstly, we construct a KD-tree to efficiently calculate the $k$-nearest neighbors of $t_i$ and sort them based on their distances from $t_i$. Next, we select the median distance, denoted as $m_i$, from this set of candidate distances as the density value of $t_i$. Using this approach, we can determine the value of $\rho$:

\begin{equation}
\rho = \frac{1}{n_t}\sum_{i=0}^{n_t}m_i, i\in[0,n_t]
\label{equ::point_cloud_density}
\end{equation}
where $n_t$ refers to the number of $t_i$, which is also equal to $n_\mathcal{C}$. To calculate the value of $r$, we use the following formula:

\begin{equation}
r=\frac{1}{\rho}
\label{equ::resolution_calculation}
\end{equation}

\subsubsection{Pixel Values Calculation}
\label{sec::pixel_values_calculation}

We construct a relative height map $I(x,y)$, where $(x,y)\in[0,r)\times[0,r)$. In this map, each pixel represents either a hole or records the relative signed height from surface $\mathcal{S}$ to point cloud $\mathcal{C}$. This construction enables us to perform quantitative processing on the high-frequency components $H$. Pixels that do not correspond to any $t_i$ are considered as holes, denoted as $\Omega$, while for the non-hole region $\Phi$, we determine their pixel values based on $D_{P_\phi}$, where $P_\phi\in\Phi$. \par

To effectively highlight the local features of point clouds, we propose a strategy that selects the extremum value from $D_{p_i}$ for each pixel. This strategy can be expressed as follows: 

\begin{equation}
    I(x,y) = {
    \begin{cases}
        Hole & {N_{D^+} + N_{D^-} = 0} \\
        \max\{D_{P_(x,y)}\} & {N_{D^+} >= N_{D^-}} \\
        \min\{D_{P_(x,y)}\} & {N_{D^+} < N_{D^-}}
    \end{cases}}
    \label{equ::height_map_pixel_value}
\end{equation}

where $N_{D^+}$ and $N_{D^-}$ represent the number of positive and non-positive values of $D_{P_{x,y}}$ in the pixel $I(x,y)$ respectively. If $I(x,y)\in\Phi$, we compare $N_{D^+}$ and $N_{D^-}$ within $I(x,y)$ and select the value of $D_{P_{x,y}}$ with the highest magnitude if $N_{D^+}$ is greater. Conversely, we choose the value with the smallest magnitude of $D_{P_{x,y}}$. \par

\subsection{Height Map Inpainting}
\label{sec::height_map_inpainting}

After representing the high-frequency components using a height map, our objective shifts to inpainting the height map. This is achieved by solving the Poisson equation, guided by patch matching in the gradient domain. \par

\begin{figure} [H]
    \centering
    \subfigure[Sampling]{
    \includegraphics[height=0.3\linewidth]{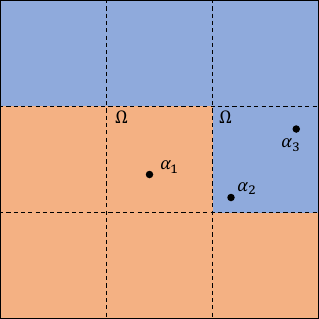}}
    \label{Fig::reconstruction_sample}
    \subfigure[Generating new points]{
    \includegraphics[height=0.27\linewidth]{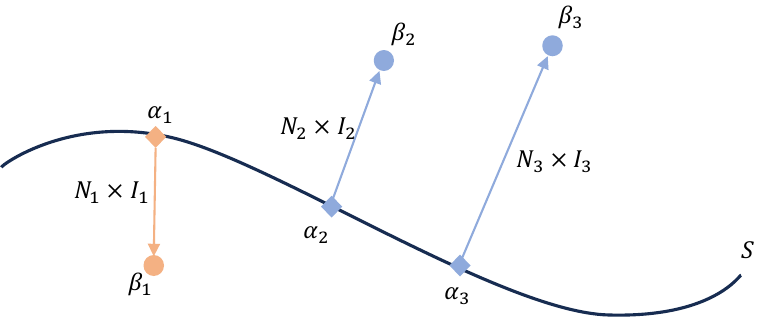}}
    \label{Fig::reconstruction_generate_new_points}
    \caption{Illustration of the reconstruction process. (\textbf{a}) We generate a set of sample points denoted as $\alpha$ within the region $\Omega$, where blue pixels represent positive intensity and orange pixels represent negative intensity. (\textbf{b}) These sample points are mapped to $\mathcal{S}$, and the intensity values $I$ are offset along their respective normals. It is important to note that if the intensity is negative, the offset direction is reversed to ensure accurate displacement. New points are denoted as $\beta$. } \label{Fig::reconstruction}
\end{figure}

\subsubsection{Constructing Poisson Equation}
\label{sec::constructing_poisson_equation}

After constructing $L$ and $H$ respectively, we transfer the task of point cloud inpainting to that of $H$ inpainting represented by a height map. We begin with an intensity image $I(x,y)$ and the desired gradients $G_x(x, y)$ and $G_y(x, y)$ inside the hole $\Omega$. We aim to inpaint a new intensity $I^*(x,y)$ within $\Omega$, subject to the constraint that $I$ and $I^*$ agree on the hole boundary $\partial\Omega$. In other words, the best $I^*(x,y)$ of $\Omega$ is the solution of the minimization problem:

\begin{equation}
\begin{aligned}
\mathop{\min}_{I^*(x,y)\in\Omega} \iint_{\Omega} {\lvert I^*(x,y) - G(x,y) \rvert}^2 \\
{\rm with}~I^*(x,y)\lvert_{\partial\Omega} = I(x,y)\lvert_{\partial\Omega}
\label{equ::poisson}
\end{aligned}
\end{equation}
The minimizer must satisfy the associated Euler-Lagrange equation, and it is the unique solution of the following Poisson equation with Dirichlet boundary conditions: 

\begin{equation}
\begin{aligned}
\Delta I^*(x,y) = {\rm div}(G_x(x,y),G_y(x,y)) \\
{\rm with}~I^*(x,y)\lvert_{\partial\Omega} = I(x,y)\lvert_{\partial\Omega}
\label{equ::poisson_solution}
\end{aligned}
\end{equation}
where $\Delta$ is the Laplacian operator and ${\rm div}$ represents the divergence of a 2D guidance field. \par

\subsubsection{Patch-match Based Inpainting in Gradient Domain}
\label{sec::patch-match_based_inpainting_in_gradient_domain}

The objective is to obtain inpainted values $G_x(x, y)$ and $G_y(x, y)$ within the hole region $\Omega$, which will solve Equation~(\ref{equ::poisson}). Typically, a well-structured patch is present in a known region of $I_\Phi$, and we employ the patch-match method to find the optimal solution for $G_x(x, y)$ and $G_y(x, y)$. \par

The core of the patch-match system revolves around the computation of patch correspondences. A Nearest-Neighbor Field (NNF) is a function $f:I\rightarrow\mathbb{R}^2$ that maps offsets to all possible patch coordinates in image $I$, utilizing some distance function between two patches. In this paper, we adopt the framework introduced by Barnes \textit{et al.}~\cite{barnes2009patchmatch} to construct all NNFs of $I_\Omega$. For each pixel $I_\Omega(x,y)$, we gather all NNFs that contain it, denoted as $F_{x,y}$. By utilizing these NNFs, we can calculate the final $G_x(x, y)$ and $G_y(x, y)$. \par

\begin{equation}
\begin{aligned}
G(x,y) = \iint_{F_{x,y}} G_{F}(x,y) \\
{\rm with}~G_{F}(x,y)\in F_{x,y}
\label{equ::calculate_gradient_pixel_value}
\end{aligned}
\end{equation}

\subsubsection{Solving Intensity Image}
\label{sec::solving_intensity_image}

After obtaining the values of all pixels belonging to $\Omega$ in the gradient domain, the next step is to solve for the true values of $I_\Omega$. We can use the known values of the pixels outside the hole pixels as boundary conditions for the reconstruction problem. The formula for this reconstruction is as follows: \par

\begin{equation}
\begin{aligned}
\Delta I^*(x,y) = \frac{\partial G_x}{\partial x}(x,y) + \frac{\partial G_y}{\partial y}(x,y) \\
{\rm with}~I^*(x,y)\lvert_{\partial\Omega} = I(x,y)\lvert_{\partial\Omega}
\label{equ::poisson_solution_for_intensity}
\end{aligned}
\end{equation}
Then we can get discrete applications in the image domain: \par

\begin{equation}
\begin{aligned}
I^*(x+1,y)+I^*(x-1,y)+I^*(x,y+1)+I^*(x,y-1) \\
-4I^*(x,y)=\frac{\partial G_x(x,y)}{\partial x} + \frac{\partial G_y(x,y)}{\partial y} \\
\label{equ::discrete_poisson_solution_for_intensity}
\end{aligned}
\end{equation}

For each pixel in the domain $\Omega$, we can obtain a matrix equation of the form $AX=B$, where $A$ represents the Laplacian operator of the image, $B$ represents the divergence operator, and $X$ represents the unknown $I^*(x,y)$. We can solve for all $I^*(x,y)$ by setting up these equations and solving a sparse linear system of them. \par

\section{Reconstruction From Decomposed Signals}
\label{sec::reconstruction_from_decomposed_signals}

As shown in Figure~\ref{Fig::reconstruction}, once all values of $I^*(x,y)$ within the hole region $\Omega$ have been solved, we can proceed to reconstruct the point cloud signals using the complete signals $H$ and $L$. Specifically, we sample points in the two-dimensional $UV$ space, which is represented as the image in Section~\ref{sec::pixel_values_calculation}. These sample points, denoted as $\alpha$, are then mapped to the three-dimensional point cloud space using Equation~(\ref{equ::b-spline_surface}). By calculating the normals and intensities of $\alpha$, we generate new points by applying offsets to them within the three-dimensional space. \par

To achieve a random distribution of $\alpha$ in $UV$ space, satisfying $\alpha \in [0,1]^2$, we utilize the Halton sequence~\cite{halton1964algorithm} to generate sampling points within $I^*(x,y)$. Each sampling point can determine its corresponding coordinates on $\mathcal{S}(u_\alpha,v_\alpha)$ in three-dimensional space. Furthermore, the partial derivatives with respect to $u$ and $v$ represent the tangent vectors at $\mathcal{S}(u,v)$:

\begin{equation}
\begin{aligned}
    \frac{\partial \mathcal{S}}{\partial u} = (\frac{\partial B_x}{\partial u},\frac{\partial B_y}{\partial u},\frac{\partial B_z}{\partial u}) \\
    \frac{\partial \mathcal{S}}{\partial v} = (\frac{\partial B_x}{\partial v},\frac{\partial B_y}{\partial v},\frac{\partial B_z}{\partial v})
\end{aligned}
\label{equ::surface_tangent_vector}
\end{equation}
In Equation~(\ref{equ::surface_tangent_vector}), the first one represents the tangent vector in the $u$-direction, while the second one represents the tangent vector in the $v$-direction. The normal vector $\vec{n}_{u,v}$ at $\mathcal{S}(u,v)$, is obtained by taking the cross-product of these partial derivatives, following the right-handed rule:

\begin{equation}
    \vec{n}_{u,v} = \frac{ {\frac{\partial \mathcal{S}}{\partial u} \times \frac{\partial \mathcal{S}}{\partial v}} } { \lvert{\frac{\partial \mathcal{S}}{\partial u} \times \frac{\partial \mathcal{S}}{\partial v}} \rvert}
\end{equation}
Additionally, the intensity values can be obtained by performing bilinear interpolation on neighboring pixels of the sample points. This enables a smoother reconstruction of the point cloud, resulting in improved coherence between the reconstructed point cloud and the original point cloud. Furthermore, we can obtain the normal vector $\vec{n}_{u,v}$ and intensity value $I_{u,v}$ through bilinear interpolation. By utilizing this method, we can successfully reconstruct the final point clouds:

\begin{equation}
\mathcal{C} = \mathcal{S}(u,v) + \Vec{n}_{u,v} \times I_{u,v};
\label{equ::reconstruct_point_clouds}
\end{equation}

\section{Handling Large-Scale Point Clouds}
\label{sec::handling_large-scale_point_clouds}

In order to effectively handle Large-scale data of terrain point cloud, we employ a data preprocessing step prior to implementing our algorithmic framework. Due to the large number of points present in raw acquired data, directly fitting it to a B-spline surface would be time-consuming. To enhance the efficiency of the fitting process, we propose a data preprocessing step in which we downsample $\mathcal{C}$ using voxelization techniques. The downsampling scheme involves spatial discretization and gravity center extraction, as demonstrated below:
\begin{enumerate}
    \item \textbf{Spatial discretization:} In this step, $\mathcal{C}$ is split into voxels. Each voxel comprises a set of points $\{\hat{p}_0, \hat{p}_1, \dots, \hat{p}_{N_v}\}$ ($\hat{p}_i\in \mathbb{R}^3$), where $\hat{p}_i$ represents the coordinates of the $i$-th point in the voxel and $N_v$ denotes the number of points in the voxel. 
    \item \textbf{Gravity center extraction:} For each non-empty voxel, we replace points inside this voxel with the gravity center $p^{\prime}$ as shown in Equation~(\ref{equ::voxelization}):

    \begin{equation}
        p^{\prime} = \frac{1}{N_v}\sum_{i=0}^{N_v}{\hat{p}_i}
        \label{equ::voxelization}
    \end{equation}
    
\end{enumerate}

The downsampled terrain point cloud serves as the input for Section~\ref{sec::low-frequency_component_inpainting}. \par

\section{Experimental Results}
\label{sec::experiments}

In this section, we demonstrate the effectiveness of our algorithm through three sets of experiments. Firstly, section~\ref{sec::results_on_low-frequency_component_inpainting} illustrates that the low-frequency component generated can effectively represent the basic shape of the input model. Section~\ref{sec::results_on_high-frequency_component_inpainting} highlights the necessity and effectiveness of the high-frequency component in our proposed point cloud representation framework. Finally, in section~\ref{sec::overall_results_and_comparisons}, we showcase our overall inpainting results and compare them with other algorithms. \par

\subsection{Experimental Setup}
\label{sec::experimental_setup}

We evaluate the proposed method by testing it on seven real-world terrain scenes, coming from NatureManufacture~\cite{naturemanufacture} and JianShan~\cite{JianShan}. The holes in these scenes are introduced by: (1) the removal of unwanted objects such as trees and vehicles, and (2) deliberately creating holes on the terrain so that we can compare the hole filling results with ground truth. In Section~\ref{sec::overall_results_and_comparisons}, each of the terrain datasets with ground truth contains approximately 100,000 point clouds, while the terrain datasets without ground truth consists of around 1 million point clouds. Additionally, it is important to note that our focus solely lies on the positional information of point clouds, disregarding their colors. Consequently, the chosen terrain data uniformly considers elevation as the color component. \par

In our experiments, we construct $L$ by empirically using a 4th-order (set $p=3$ and $q=3$) B-spline surface to ensure that the fitting result neither incurs excessive cost nor loses significant details. Increasing the number of control points can make $L$ more refined, but this also implies higher costs. As a trade-off between accuracy and efficiency, we set both $m$ and $n$ as 19. Thus, the uniformly spaced knot vectors $u = (u_0,u_1,\dots,u_{22})$, $v = (v_0,v_1,\dots,v_{22})$ can be obtained where $u_i = \frac{i}{22}$ and $v_j = \frac{j}{22}$ with $i,j\in[0,22]$. The iteration size to perform fitting optimization is $I=10$. Additionally, during preprocessing, we set the ratio of voxel edge length to the longest axis length of the oriented bounding box (OBB) to 0.05. For image inpainting, we utilize a patch size of 11 and conduct 10 iterations of patch-match. \par

Specifically, we compare our methods with several established traditional and learning-based methods, including one mesh-based traditional method Attene \textit{et al.}~\cite{attene2010lightweight} (MeshFix), three point-based traditional methods, i.e. Doria \textit{et al.}~\cite{doria2012filling} (DGI), Altantsetseg \textit{et al.}~\cite{altantsetseg2017complex} (CSF), Shi \textit{et al.}~\cite{shi2022point} (NBFM) and three learning-based point cloud completion methods, i.e. Alliegro \textit{et al.}~\cite{alliegro2021denoise} (Deco), Xiang\textit{et al.}~\cite{xiang2021snowflakenet} (Snowflake), lin \textit{et al.}~\cite{lin2023hyperbolic} (HyperCD).

\subsection{Evaluation Metric of Inpainting Results}
\label{sec::evaluation_metric_of_inpainting_results}

We apply two evaluation metrics named GPSNR and NSHD to measure geometric differences between two point clouds objectively. \par

GPSNR measures the distortion between point cloud $A$ and $B$ with respect to a peak value: \par

\begin{equation}
 GPSNR_{A, B} = 10 \log_{10}{\frac{p_d^2}{e_{A, B}}}  \label{GPSNR}
\end{equation}
where $p_d$ refers to the diagonal distance of a bounding box of the point cloud and $e_{A, B}$ represents point-to-plane distances. Therefore a higher GPSNR means a lower gap between two point clouds. \par

NSHD (Normalized Symmetric Hausdorff Distance) is the normalized version of the One-sided Hausdorff Distance: \par

\begin{equation}
 NSHD_{A, B} = \frac{1}{V} max\{OHD_{A, B}, OHD_{B, A}\}  \label{NSHD}
\end{equation}
where $V$ denotes the volume of the oriented bounding box of a point cloud, and $OHD_{A, B}$ represents the One-sided Hausdorff Distance from point cloud A to point cloud B. Therefore, the smaller the NSHD, the more similar the two point clouds are. However, NSHD is closely related to point density, so comparisons between different test scenarios may not be meaningful. \par

\begin{figure}[h]
    \centering
    \subfigure[Input point cloud with hole]{
    \includegraphics[width=0.47\linewidth]{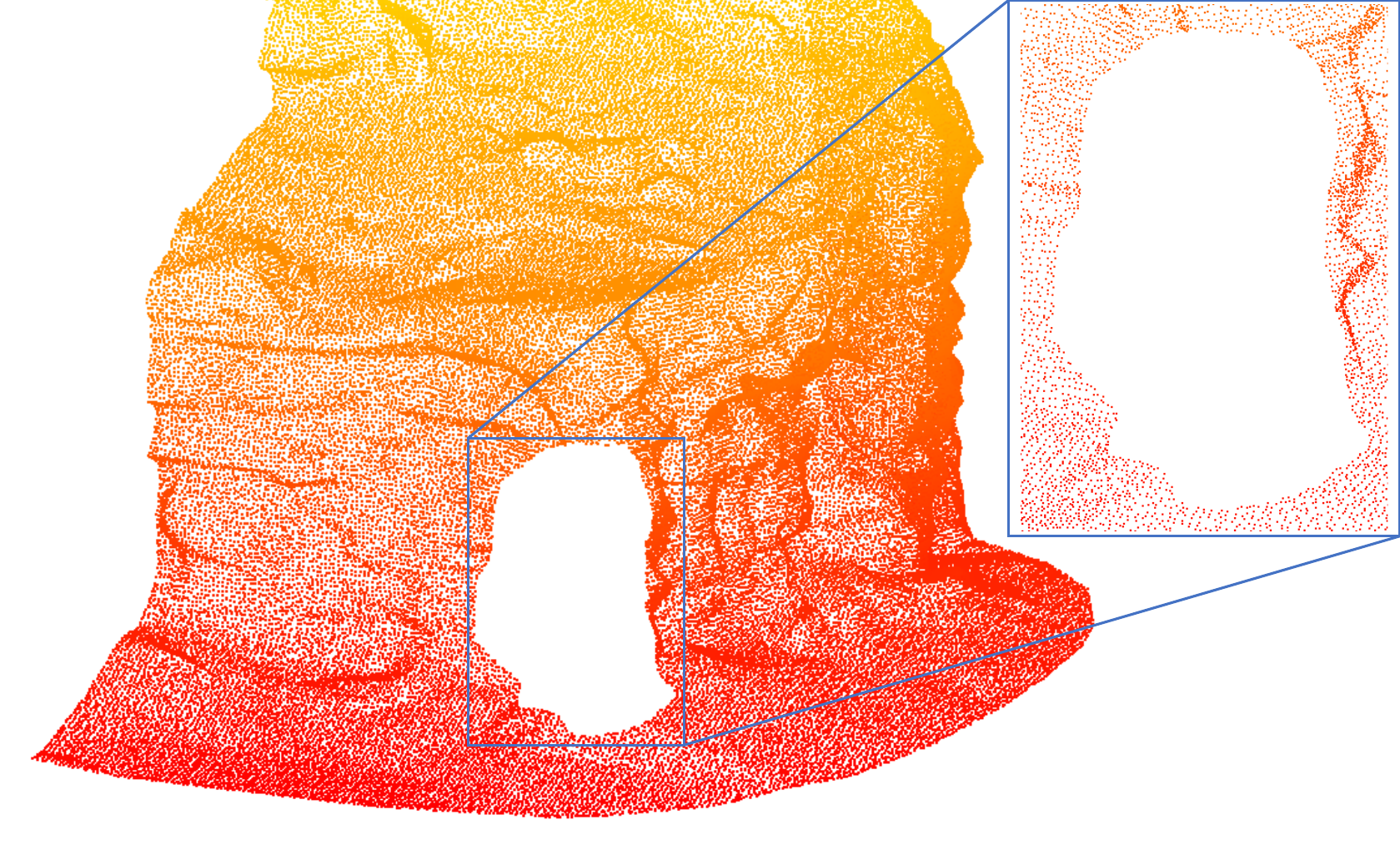}}
    \subfigure[Low-frequency component represented as B-spline surface]{
    \includegraphics[width=0.47\linewidth]{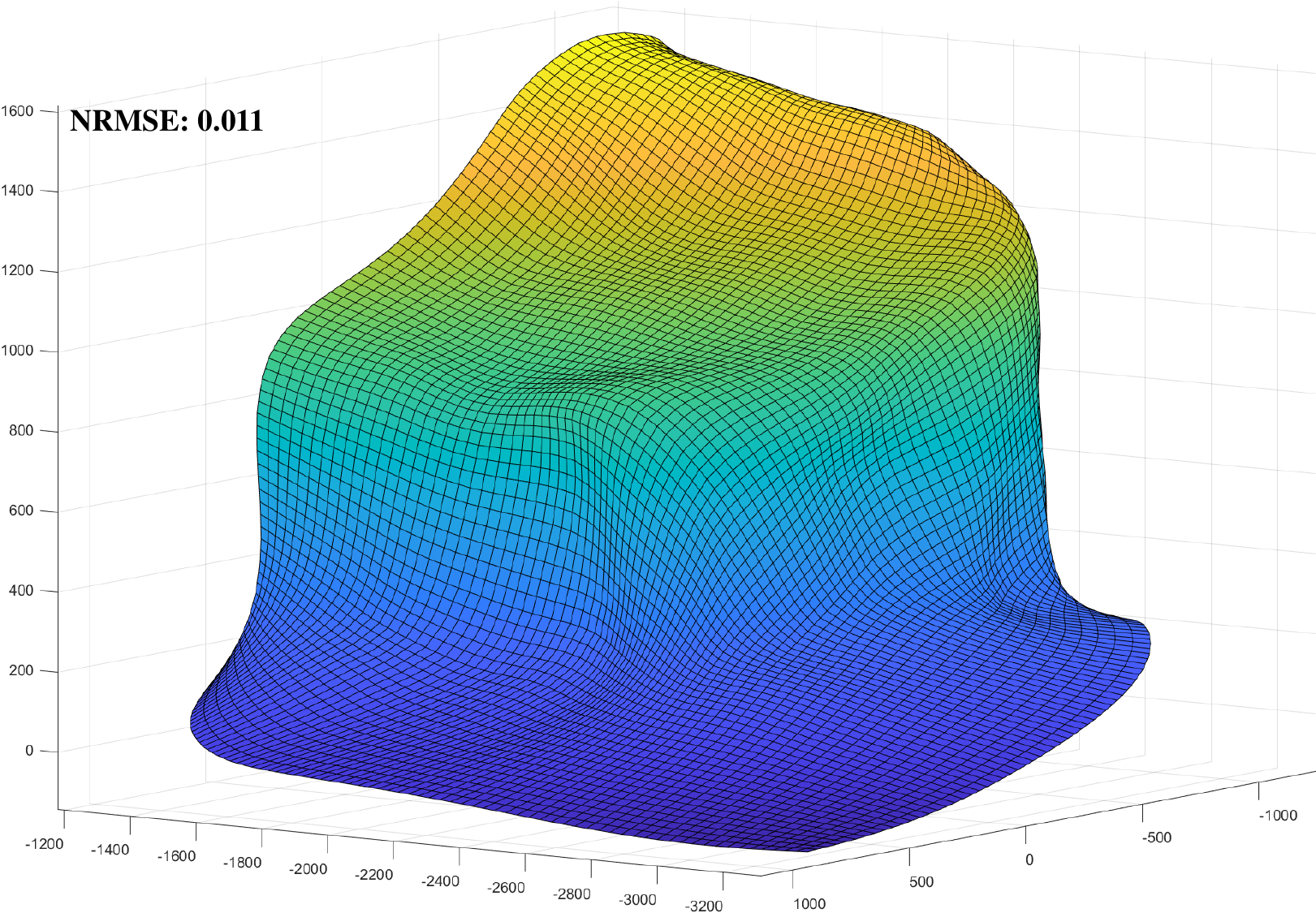}}
    \caption{Illustration of the low-frequency component result. (\textbf{a}) presents the input point cloud model, which contains holes in its structure. (\textbf{b}) showcases the fitted B-spline surface obtained through our algorithm and NRMSE value between itself and (\textbf{a}).}
    \label{Fig::low-fre-result}
\end{figure}

\subsection{Results on Low-frequency Component Inpainting}
\label{sec::results_on_low-frequency_component_inpainting}

As depicted in Figure~\ref{Fig::low-fre-result}, we present the obtained low-frequency component. From a subjective visual perspective, we leverage the flexibility and smoothness properties of the B-spline surface to accurately fit a shape that closely resembles the original point cloud model. Through this fitting process, we effectively capture the overall shape and curvature characteristics of the original point cloud, ensuring a faithful representation of its global shape and curvature features. To objectively evaluate the accuracy of the fitting results and quantify the discrepancy between the fitted surface and the original signal, we employed NRMSE (Normalized Root Mean Square Error) as the evaluation metric. NRMSE is a value that ranges from 0 to 1, with values closer to 0 indicating a higher similarity between the predictive model and the truth, while values closer to 1 suggest the opposite. In our specific case, the obtained NRMSE value of 0.011 demonstrates that our low-frequency component closely approximates the input point cloud to a significant extent. \par

\begin{figure}
    \centering
    \subfigure[Ours]{
    \includegraphics[width=0.47\linewidth]{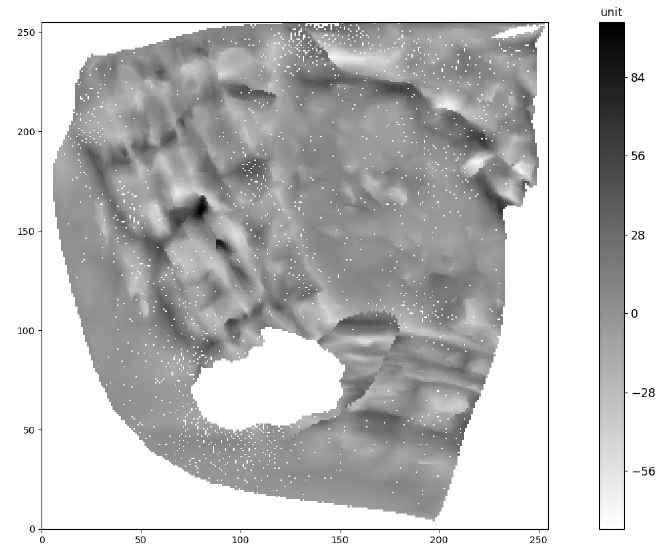}}
    \subfigure[DGI]{
    \includegraphics[width=0.47\linewidth]{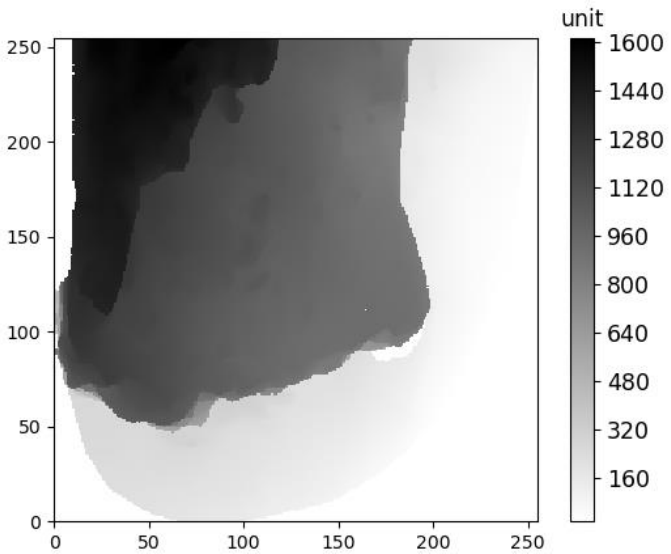}}
    \subfigure[Projection Distribution in Ours]{
    \includegraphics[width=0.47\linewidth]{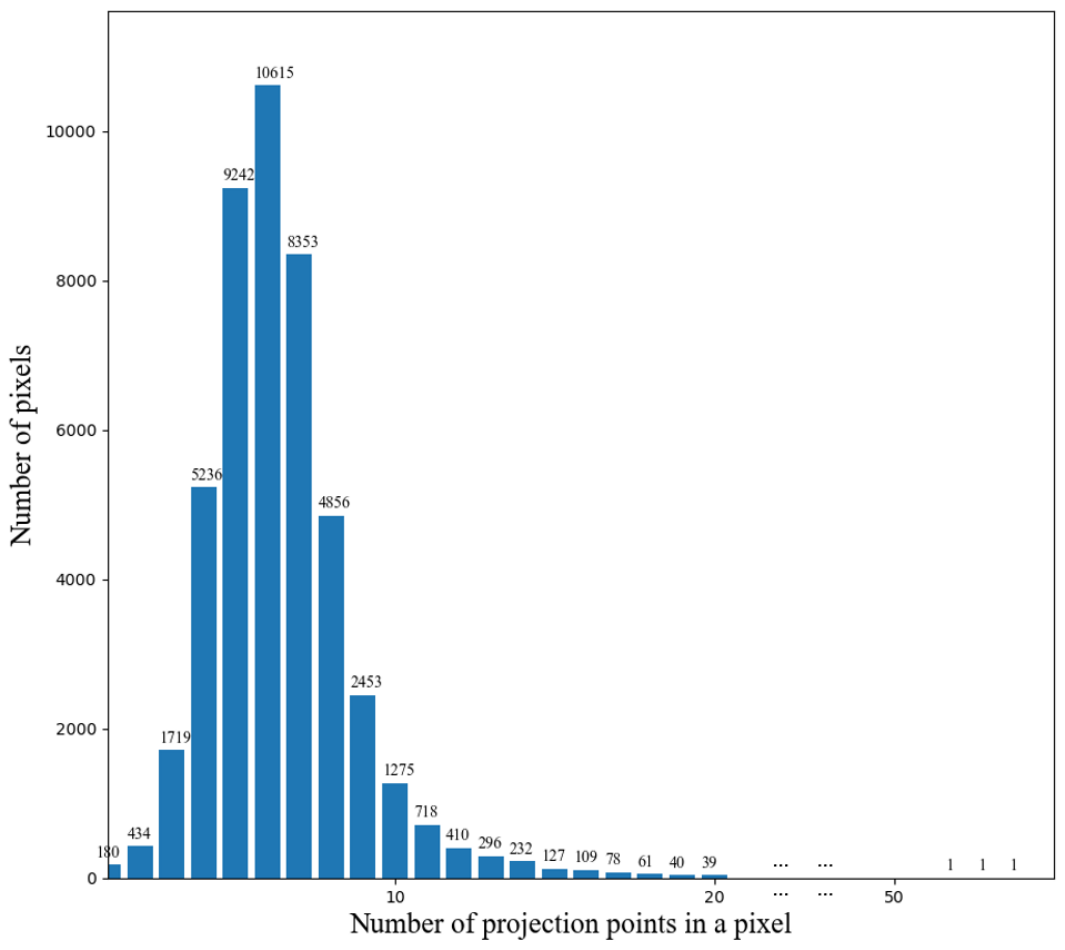}}
    \subfigure[Projection Distribution in DGI]{
    \includegraphics[width=0.47\linewidth]{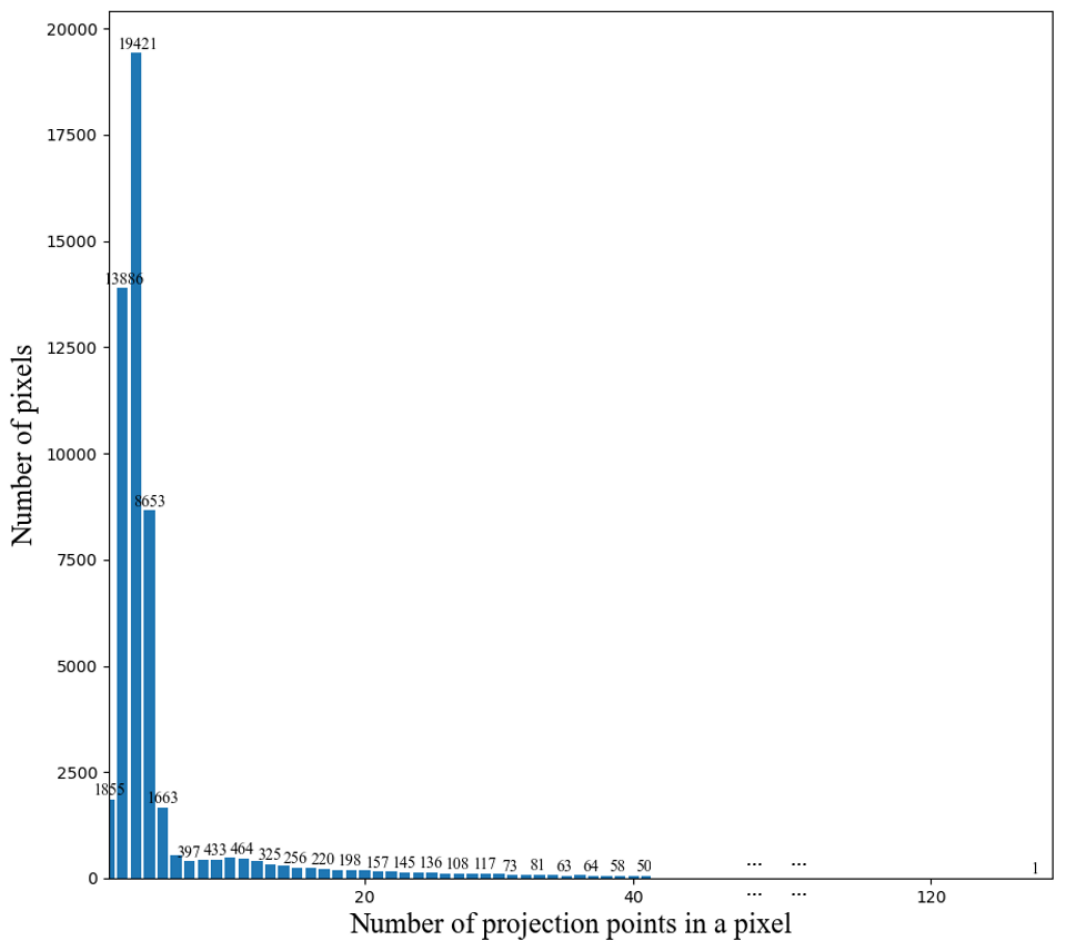}}
    \caption{(\textbf{a}) The height map based on the B-spline surface in ours. (\textbf{b}) The height map based on $XY$ plane in DGI. (\textbf{c}) Distribution of the number of projection points per pixel in Ours. (\textbf{c}) Distribution of the number of projection points per pixel in DGI.}
    \label{Fig::low-fre-height-map}
\end{figure}

DGI employs a representation similar to DEM, utilizing $XY$ plane as the low-frequency signal. However, our approach to selecting the low-frequency component is more reasonable. Figure~\ref{Fig::low-fre-height-map} illustrates this by showcasing the generation of corresponding unrepaired high-frequency components. Subjectively, our height map exhibits more natural color variations, while the height map obtained by DGI displays significant differences between neighboring pixels, indicating abrupt changes in elevation and potential loss of elevation information. Objectively, an analysis of the number of projection points per pixel in the height map reveals a more balanced distribution of the original point cloud across our generated low-frequency signal. In our algorithm, the proportion of pixels with less than 10 projection points is approximately 95$\%$, with a maximum of 54 projection points. In contrast, for DGI, these figures are 89$\%$ and 128, respectively. \par

\subsection{Results on High-frequency Component Inpainting}
\label{sec::results_on_high-frequency_component_inpainting}

\begin{figure}[h]
    \centering
    \subfigure[Hole]{
    \includegraphics[width=0.31\linewidth]{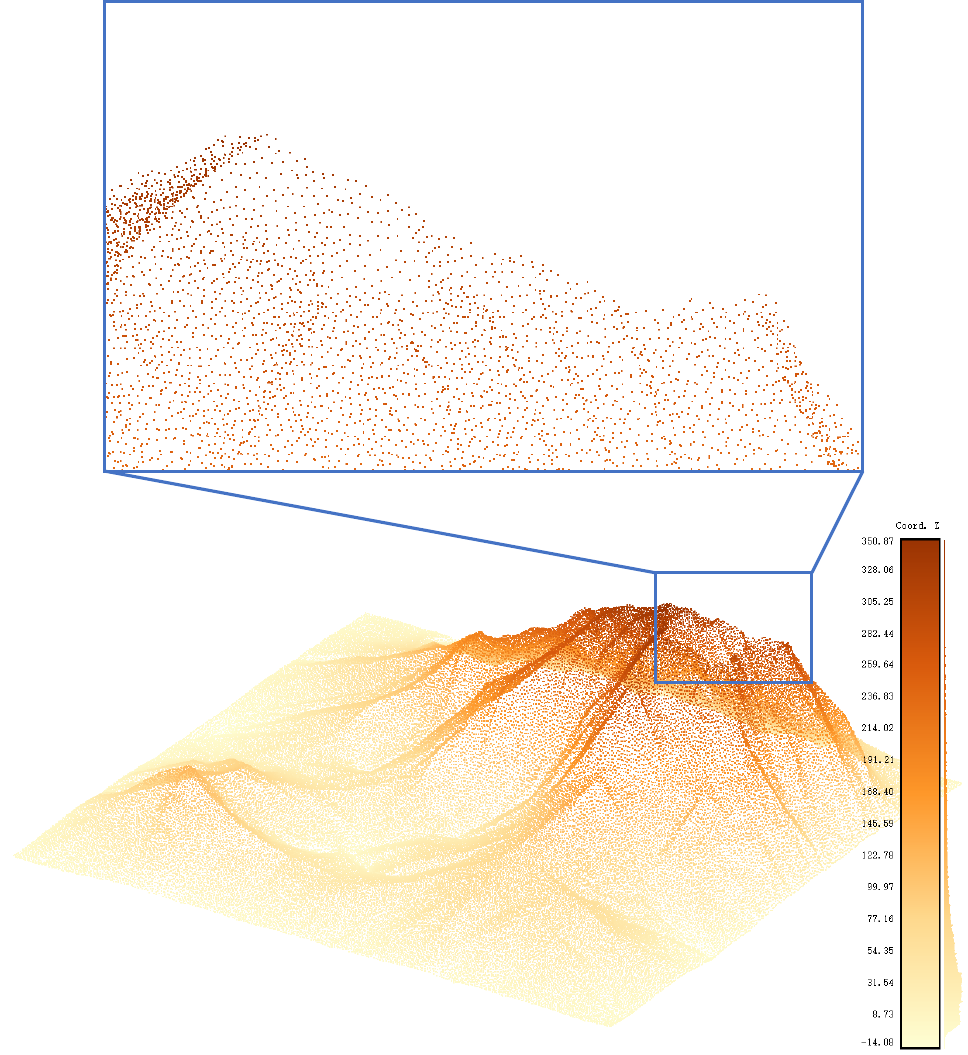}}
    \subfigure[Inpaint only by low-frequency component]{
    \includegraphics[width=0.31\linewidth]{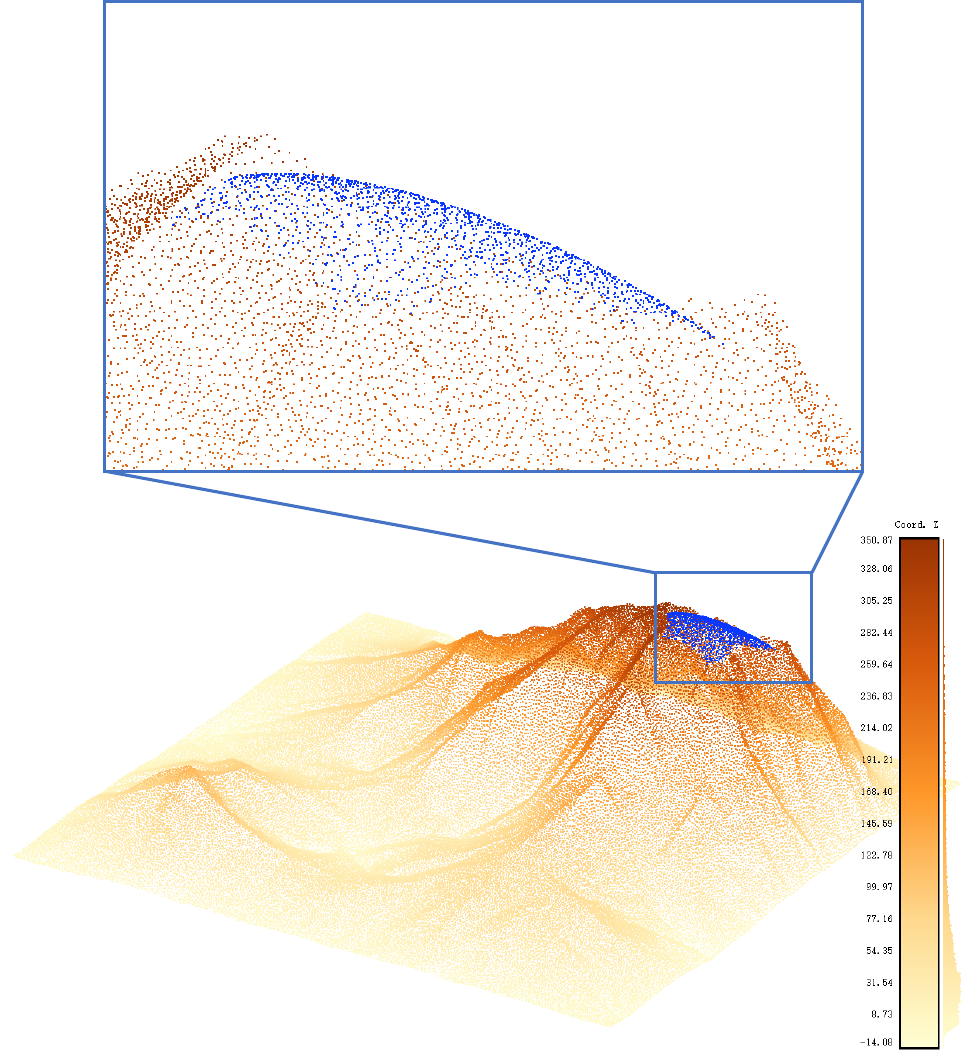}}
    \subfigure[Inpaint after reconstruction]{
    \includegraphics[width=0.31\linewidth]{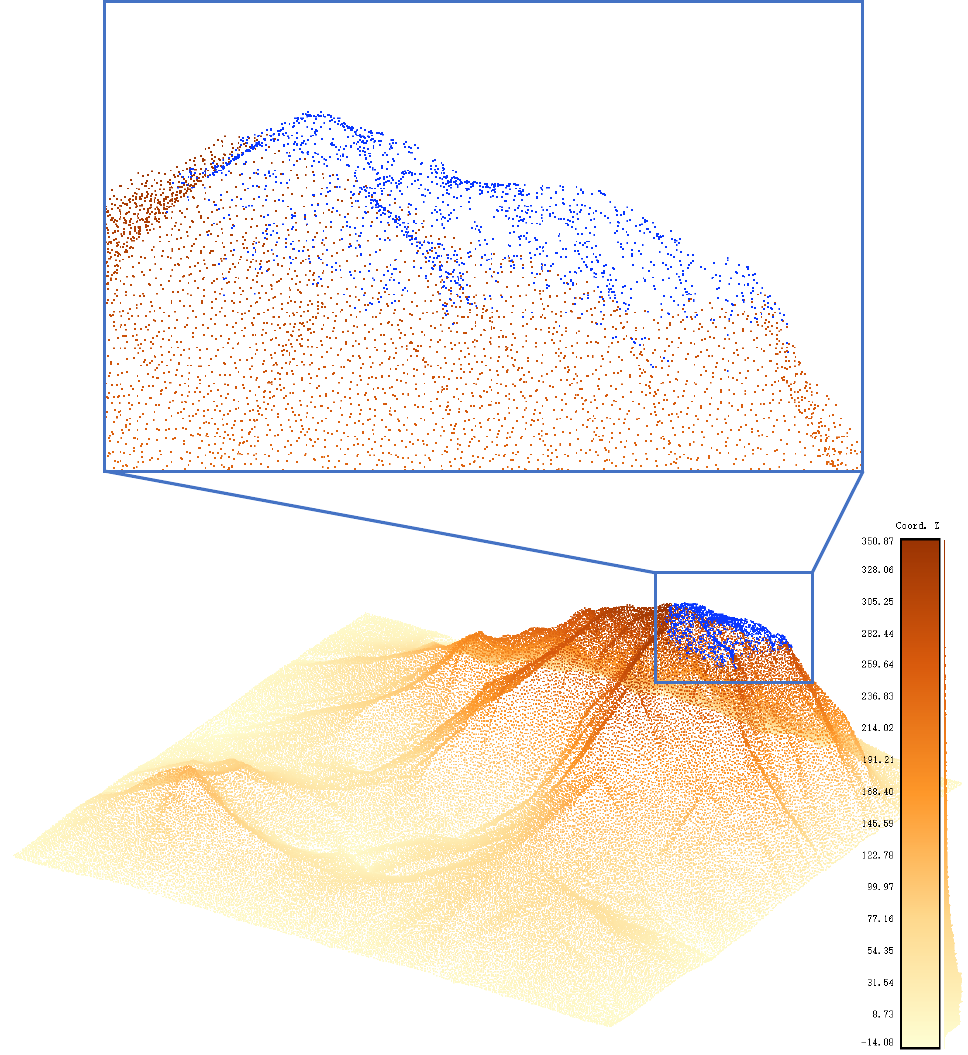}}
    \subfigure[B-spline surface]{
    \includegraphics[width=0.31\linewidth]{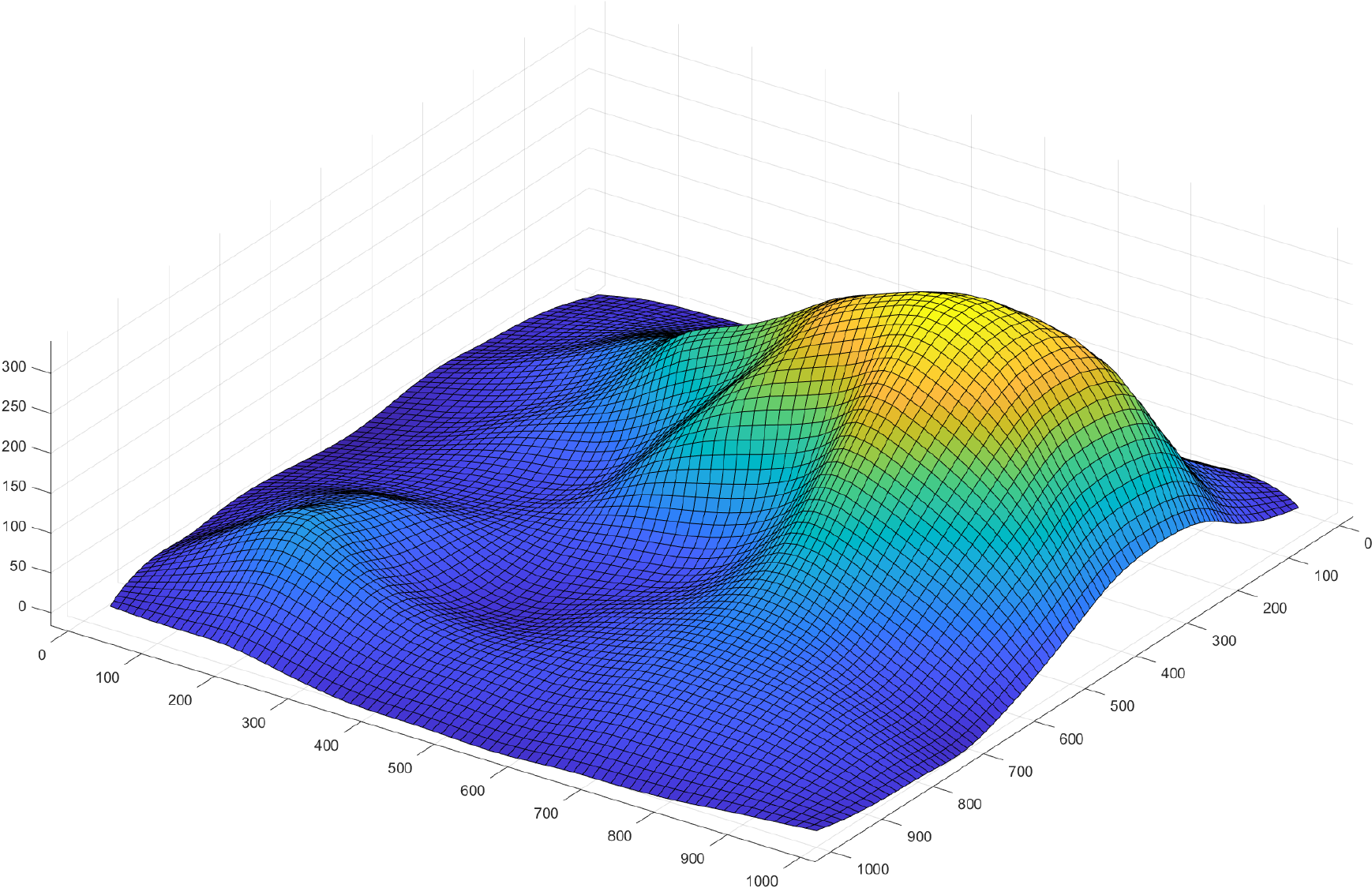}}
    \subfigure[Height map with hole]{
    \includegraphics[width=0.31\linewidth]{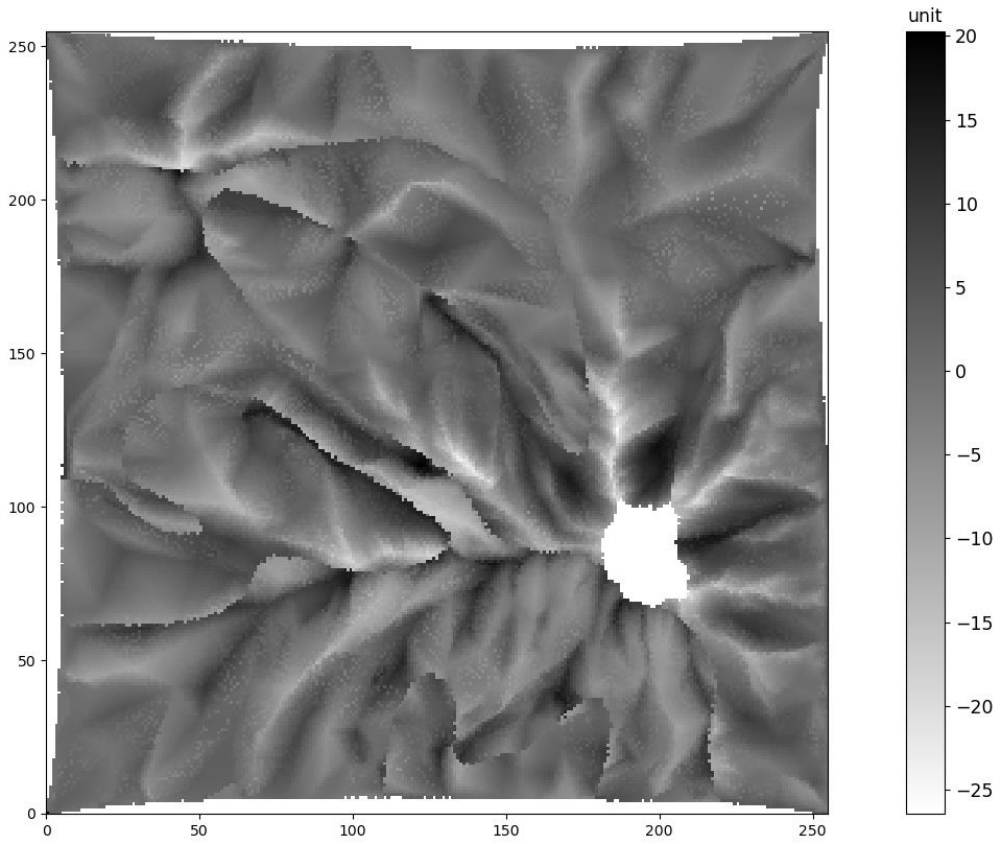}}
    \subfigure[Height map inpainted]{
    \includegraphics[width=0.31\linewidth]{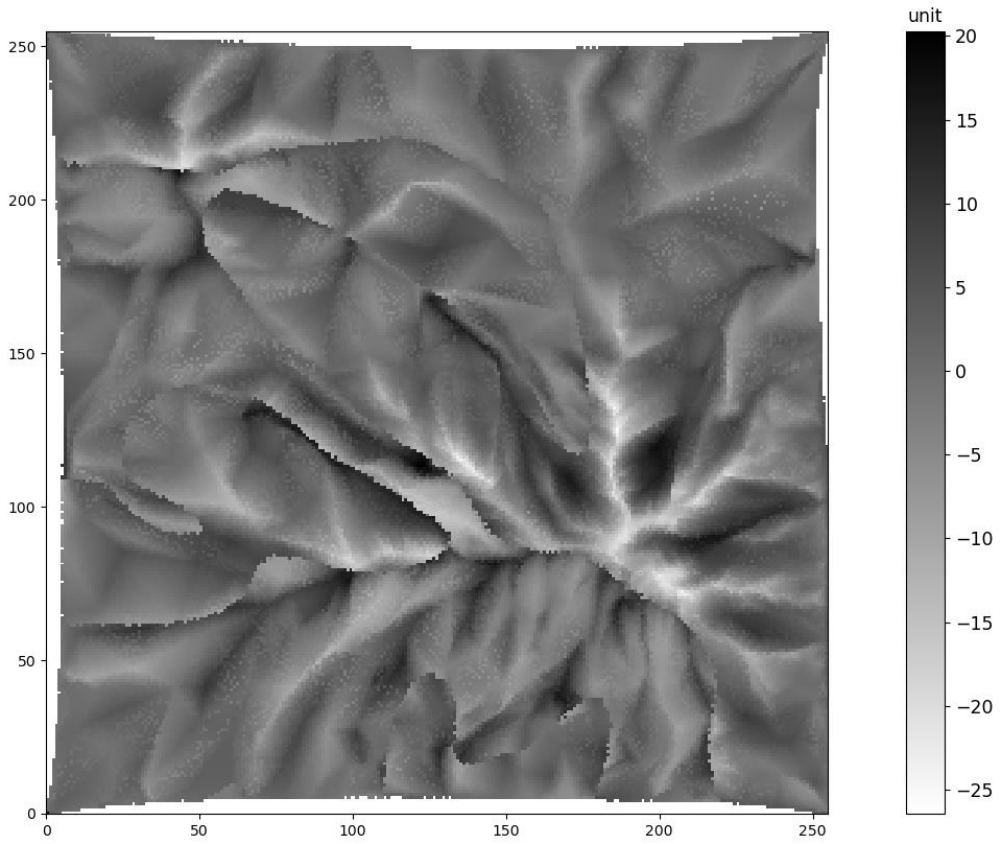}}
    \caption{Illustration of necessity of high-frequency component, in which inpainted points are marked in blue. (\textbf{a}) Original Point Cloud with Holes. (\textbf{b}) Inpaint result only using low-frequency component, which generated by sampling points on the surface. (\textbf{c}) Final result after applying both low- and high-frequency components. (\textbf{d}) Fitted B-spline surface. (\textbf{e}) Height map to be repaired. (\textbf{f}) Height map after filling holes.}
    \label{Fig::high-fre-comparison}
\end{figure}

Figure~\ref{Fig::high-fre-comparison} illustrates the comparison between the results obtained by inpainting only low-frequency component and those of a complete reconstruction, highlighting the effectiveness of high-frequency component. Subjectively, the inpainting result using only low-frequency component appears excessively smooth due to direct sampling from the fitted B-spline surface, which lacks geometric details and fails to capture the natural characteristics of the terrain. Additionally, the low-frequency surface is theoretically limited in its ability to pass through all input points due to the constraints imposed by the number of B-spline control points. Consequently, relying solely on low-frequency component leads to a lack of smooth transition with the surrounding point cloud. In contrast, incorporating both low-frequency and high-frequency components yields more natural results. By considering neighborhood information during the reconstruction of high-frequency component, the generated 3D point clouds exhibit a certain level of smoothness with the original points, without noticeable gaps along the boundary. \par

\begin{figure} [h]
    \centering
    \subfigure[Ground truth]{
    \includegraphics[width=0.31\linewidth]{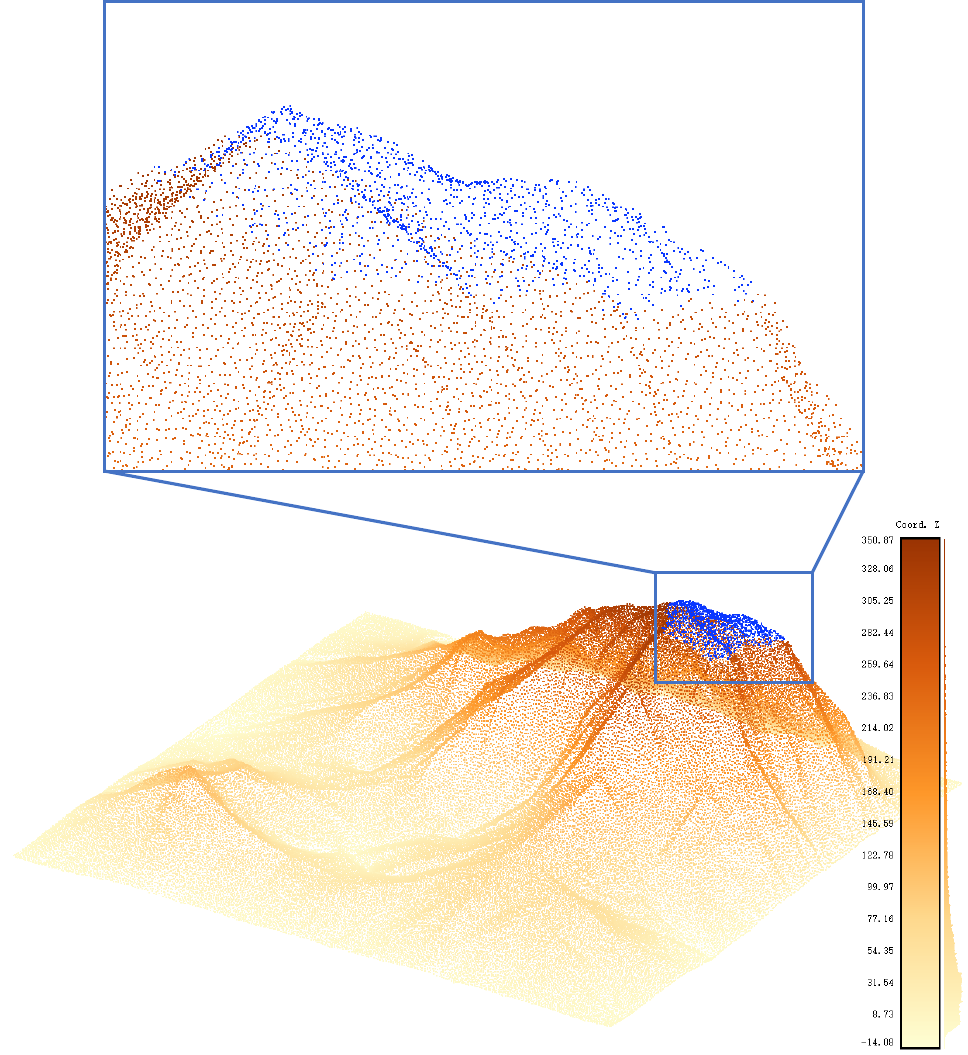}}
    \subfigure[Inpaint by our repairing scheme]{
    \includegraphics[width=0.31\linewidth]{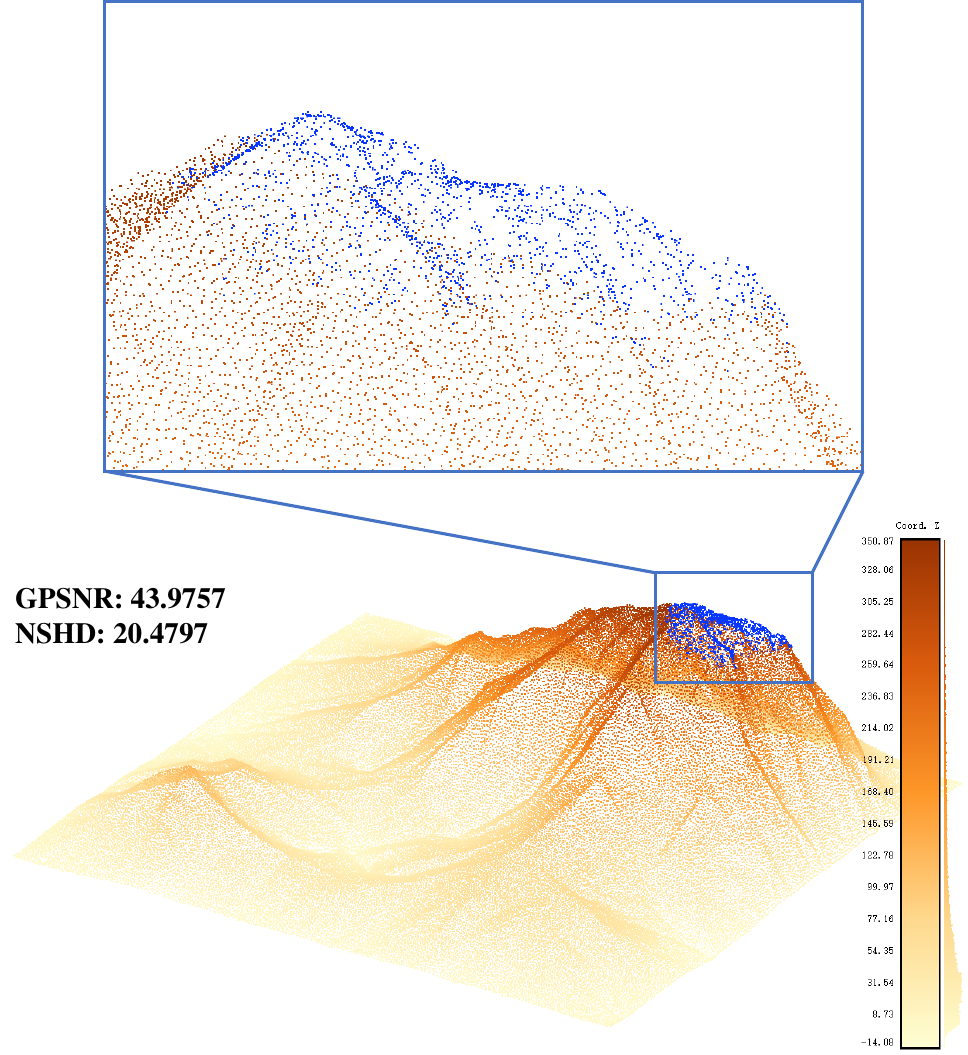}}
    \subfigure[Inpaint by noise]{
    \includegraphics[width=0.31\linewidth]{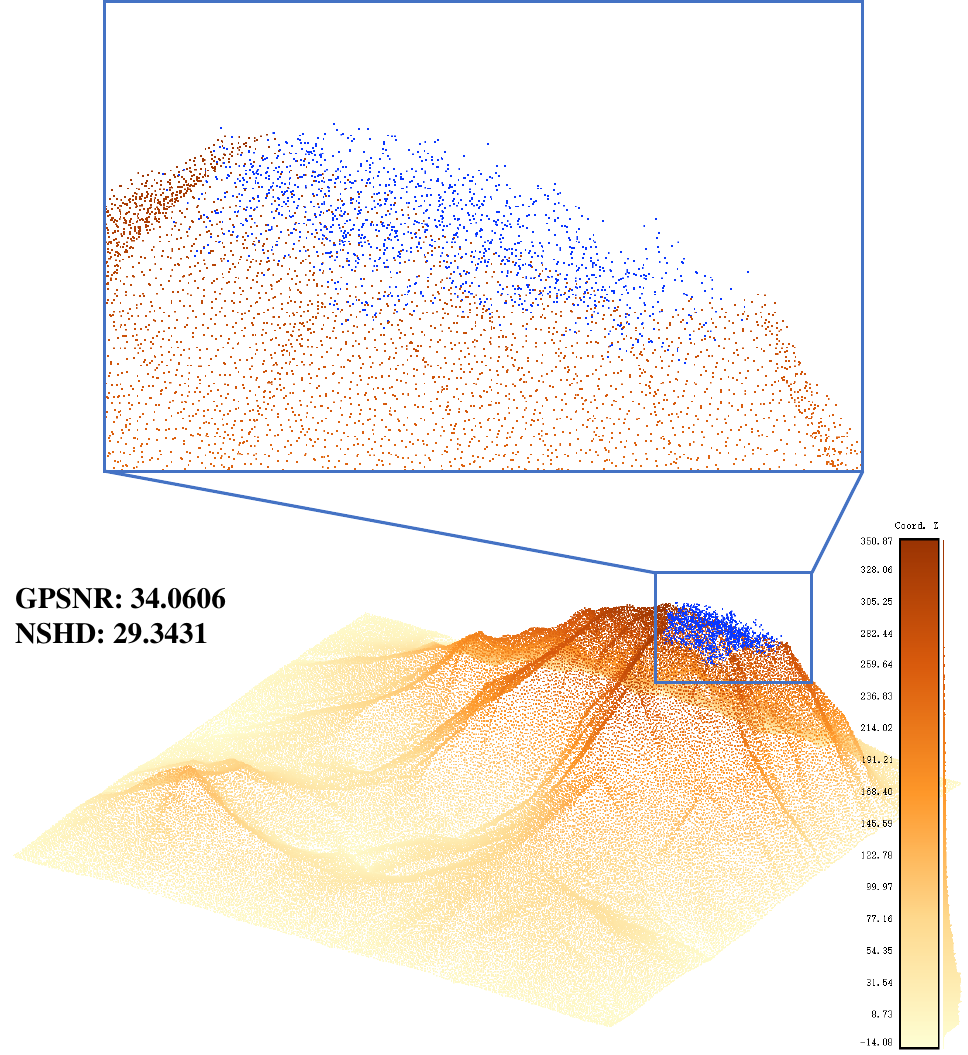}}
    \subfigure[Height map in ground truth]{
    \includegraphics[width=0.31\linewidth]{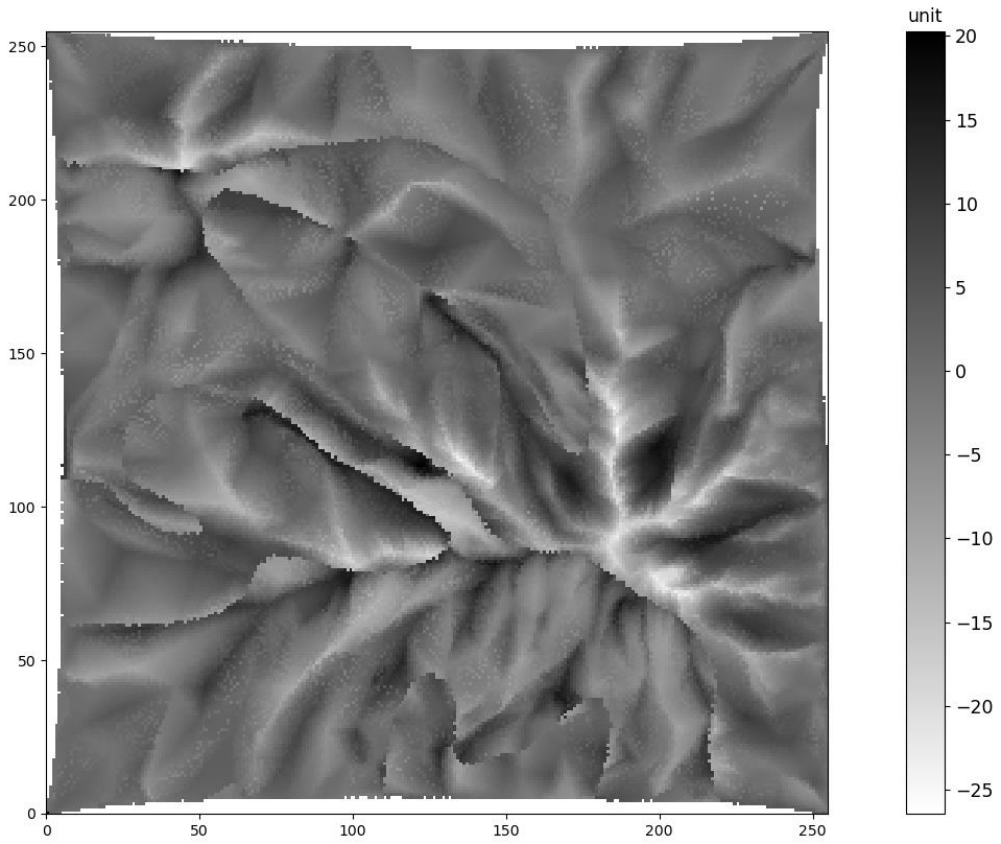}}
    \subfigure[Height map in ours]{
    \includegraphics[width=0.31\linewidth]{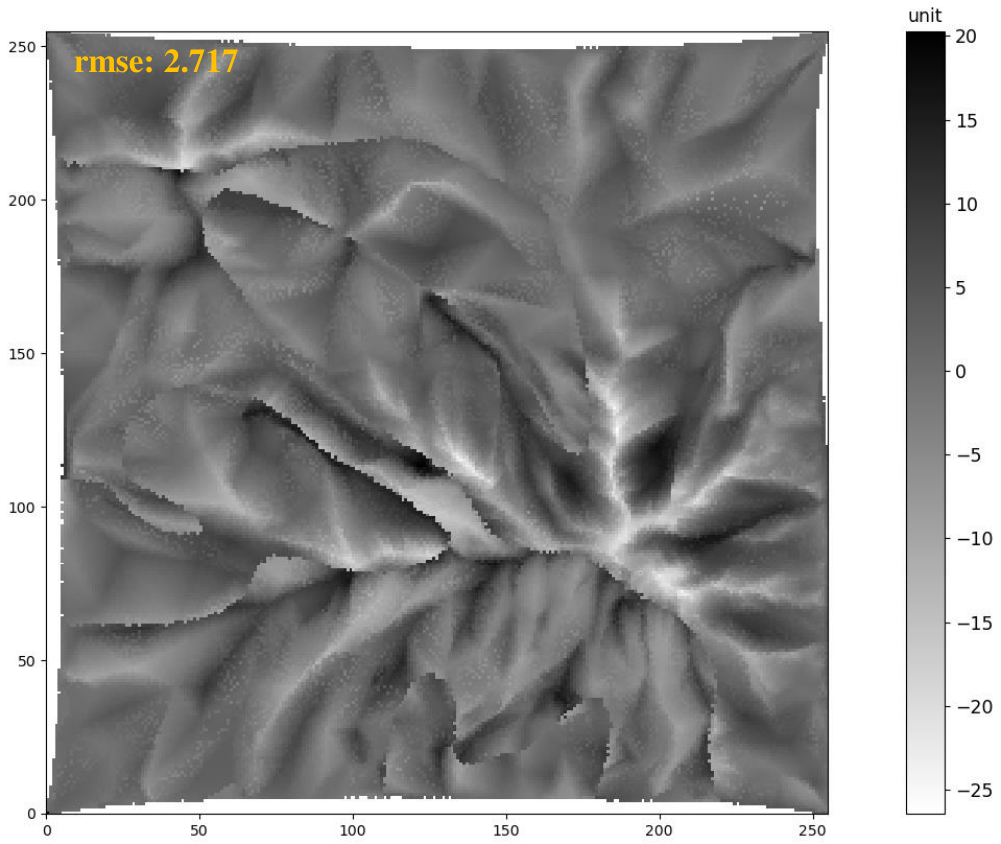}}
    \subfigure[Height map with noise]{
    \includegraphics[width=0.31\linewidth]{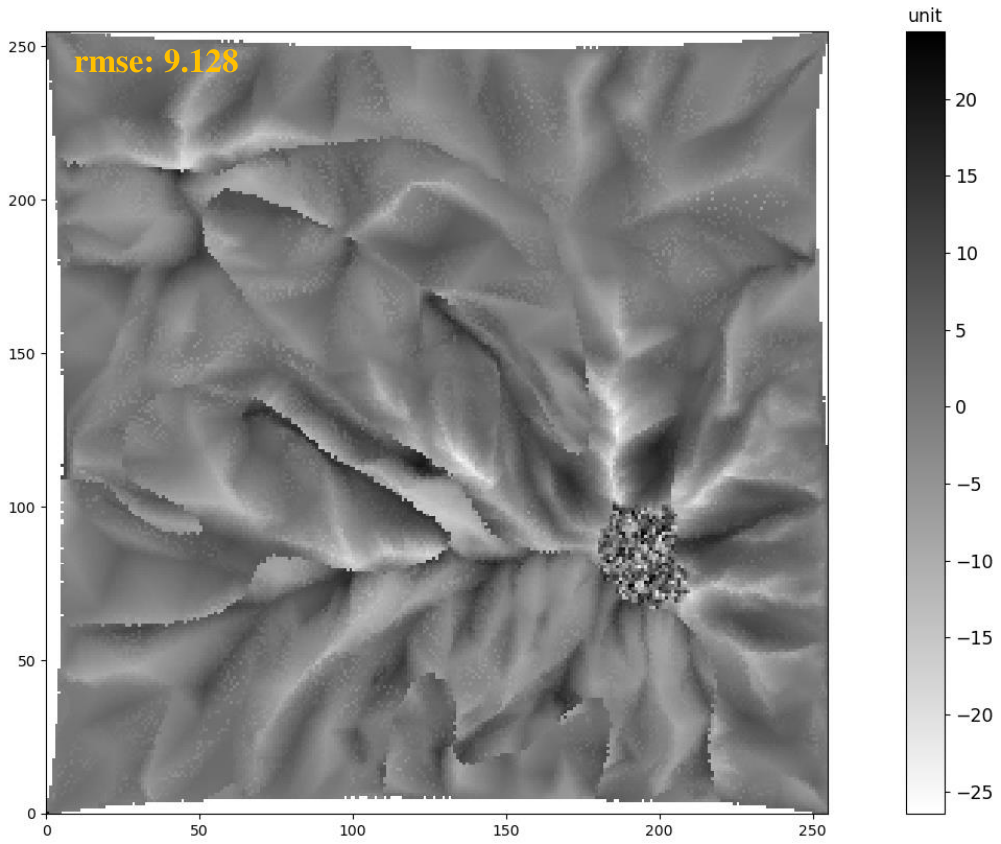}}
    \caption{Illustration of effectiveness of image inpainting in high-frequency component repairing strategy, in which a comparison is conducted between ours and random noise. (\textbf{a}) Ground truth. (\textbf{b}) Final result inpainted by our height map repairing scheme. (\textbf{c}) Final result inpainted by random noise in height map. (\textbf{d}) Ground truth's height map. (\textbf{e}) Inpainted height map in ours. (\textbf{f}) Filled height map with noise.}
    \label{Fig::high-fre-height-map-comparison}
\end{figure}

Regarding the inpainting of high-frequency component, our strategy of solving the Poisson equation guided by the gradient domain which is inpainted by patch matching, can generate results that are closer to the ground truth. Figure~\ref{Fig::high-fre-height-map-comparison} presents a comparison between our height map inpainting method and the random noise inpainting approach. From a subjective visual perspective, our approach exhibits a resemblance between the inpainted height map and that of ground truth, as well as the final reconstructed point cloud. In contrast, employing the strategy of randomly filling the holes in the height map leads to a disordered point cloud representation. Additionally, we evaluate the differences between the two height map inpainting strategies and the ground truth using RMSE, and assess the final generated point cloud using GPSNR and NSHD. Our evaluation results consistently outperform those of the random filling approach, providing objective evidence of the superior performance of our height map inpainting strategy. \par

\subsection{Overall results and comparisons}
\label{sec::overall_results_and_comparisons}

\begin{figure*} [h]
    \includegraphics[width=1\linewidth]{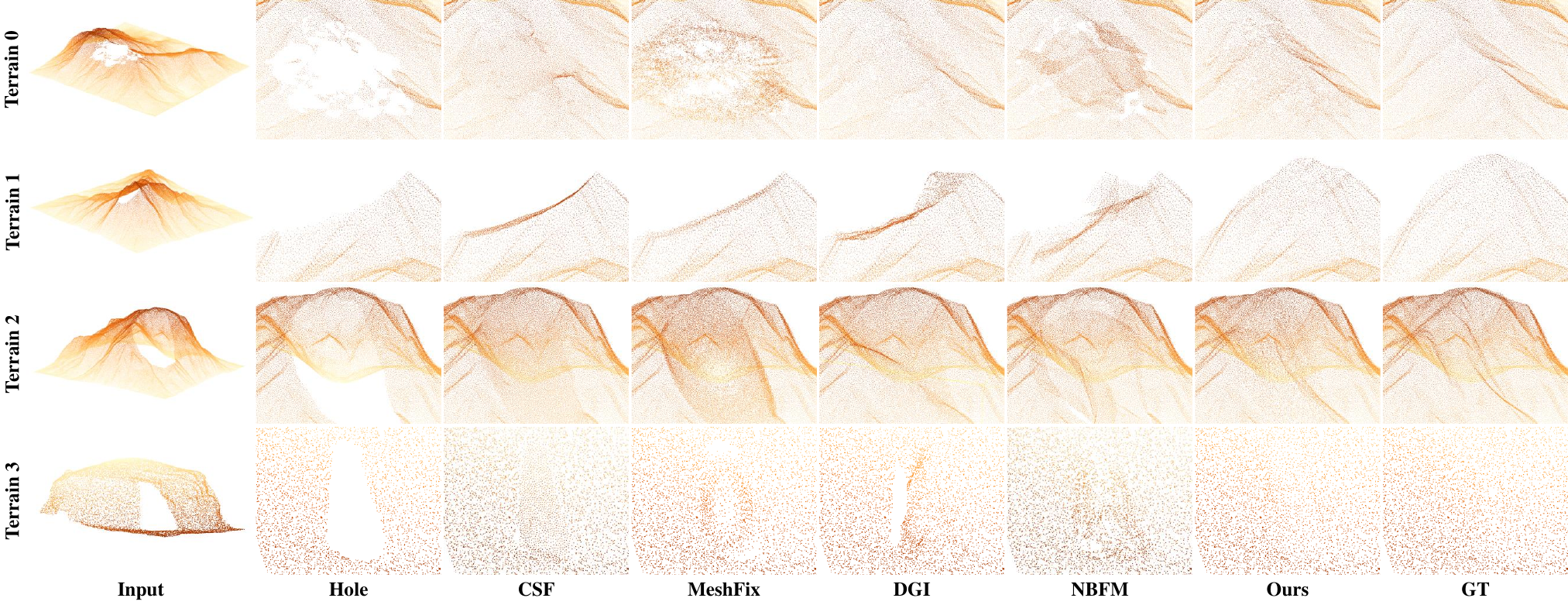}
\caption{We conducted tests on some models with ground truth available. We presented the input, ground truth, as well as the results of our algorithm and four other algorithms.}
\label{Fig::comparisons_with_others}
\end{figure*}

\begin{figure*} [h]
    \includegraphics[width=1\linewidth]{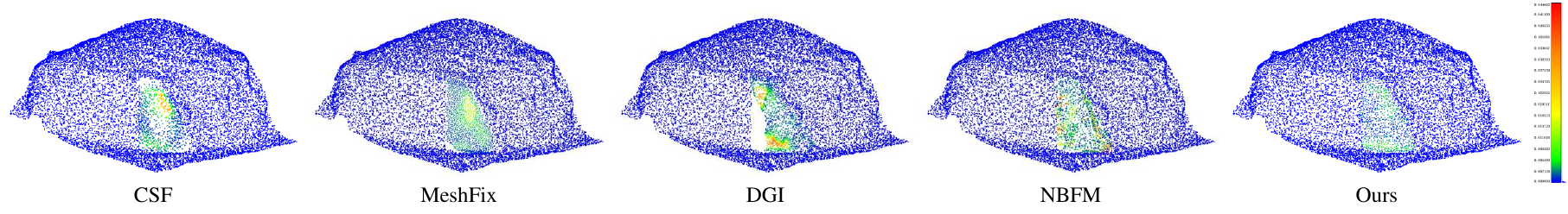}
\caption{Illustration of Chamfer Distance: The points are color-coded based on their distance to the nearest point in the ground truth point cloud. Bluer colors indicate shorter distances, while redder colors indicate longer differences.}
\label{Fig::error map}
\end{figure*}

To facilitate numerical analysis, we test on some terrain models with complex holes created artificially for the purpose of comparing our algorithm with others. For convenience, we test on several terrain point clouds, abbreviated as “Terrain”. A comparison with traditional algorithms is shown in Figure~\ref{Fig::comparisons_with_others}, and Figure~\ref{Fig::error map} demonstrates the error map to highlight the distribution of errors. From a subjective visual perspective, our algorithm performs well in generating point clouds with uniform density and natural characteristics around both complex boundary holes and no-well-defined holes. However, other algorithms encountered their problems. CSF fails to handle point cloud holes like Terrain0, which lack well-defined boundaries, due to the requirement for accurate extraction of hole boundaries and sorting of boundary points. This limitation leads to highly erroneous results. The density of points inpainted by CSF may be uneven since it necessitates iterative convergence towards the interior of the hole. For instance, in Terrain1 and Terrain2, the density at the center of the hole is lower than that at the hole's edges. On the other hand, MeshFix detects and fills holes by using mesh representation, resulting in generated models that lack geometric details. Similar to the outcomes in Terrain1 and Terrain2, it produces point cloud structures resembling cross-sections at the holes, while maintaining a very smooth appearance for inpainted points. Although DGI performs well in handling most cases, it tends to disregard information along the vertical axis, resulting in an unresolved hole region in Terrain3. Moreover, the adoption of a greedy marching strategy during the height map inpainting process leads to a noticeable lack of smoothness in the central region of holes, which is evident in Terrain1. NBFM employs a cube-based division of the point cloud space to perform inpainting. It selects appropriate cubes from the complete region of the point cloud to align with and replace the hole area. While this method effectively fills small holes, it encounters convergence challenges when dealing with large hole areas, as shown in Terrain0 and Terrain1. The difficulty lies in accurately estimating the cube size, as excessively small cubes accentuate local structural characteristics, while overly large cubes hinder point cloud alignment. Moreover, NBFM relies on boundary point detection to determine the inpainting area, resulting in the introduction of additional boundary points if the accuracy of point cloud registration is not satisfactory. \par

\begin{table}
\caption{PERFORMANCE COMPARISON IN GPSNR (DB)}
\label{Tab::GPSNR}
\setlength\tabcolsep{2pt}
\begin{tabular}{ccccccc}
\toprule
& CSF~\cite{altantsetseg2017complex} & MeshFix~\cite{attene2010lightweight} & DGI~\cite{doria2012filling} & NBFM~\cite{shi2022point} & Ours\\
\midrule
 Terrain0 & 8.3976 & 17.036 & 31.550 & 30.207 & \pmb{40.036}\\
 Terrain1 & 25.663 & 21.376 & 34.739 & 23.315 & \pmb{50.745}\\
 Terrain2 & 12.128 & 12.054 & 20.629 & 17.341 & \pmb{40.044}\\
 Terrain3 & 28.014 & 27.960 & 26.241 & 28.091 & \pmb{30.640}\\
\bottomrule
\end{tabular}
\end{table}

\begin{table}
\caption{PERFORMANCE COMPARISON IN NSHD}
\label{Tab::NSHD} 
\setlength\tabcolsep{2pt}
\begin{tabular}{ccccccc}
\toprule
& CSF~\cite{altantsetseg2017complex} & MeshFix~\cite{attene2010lightweight} & DGI~\cite{doria2012filling} & NBFM~\cite{shi2022point} & Ours\\
\midrule
 Terrain0 & 149.40	& 38.264 & 20.544 & 33.250 & \pmb{9.0525}\\
 Terrain1 & 59.728	& 74.045 & 38.221 & 56.110 & \pmb{21.319}\\
 Terrain2 & 54.491   & 66.988 & 25.993 & 48.682 & \pmb{19.510}\\
 Terrain3 & 0.3190   & 0.1310 & 0.2261 & 0.2584 & \pmb{0.1074}\\
\bottomrule
\end{tabular}
\end{table}

\begin{figure} [h]
    \centering
    \subfigure[Hole]{
    \includegraphics[width=0.47\linewidth]{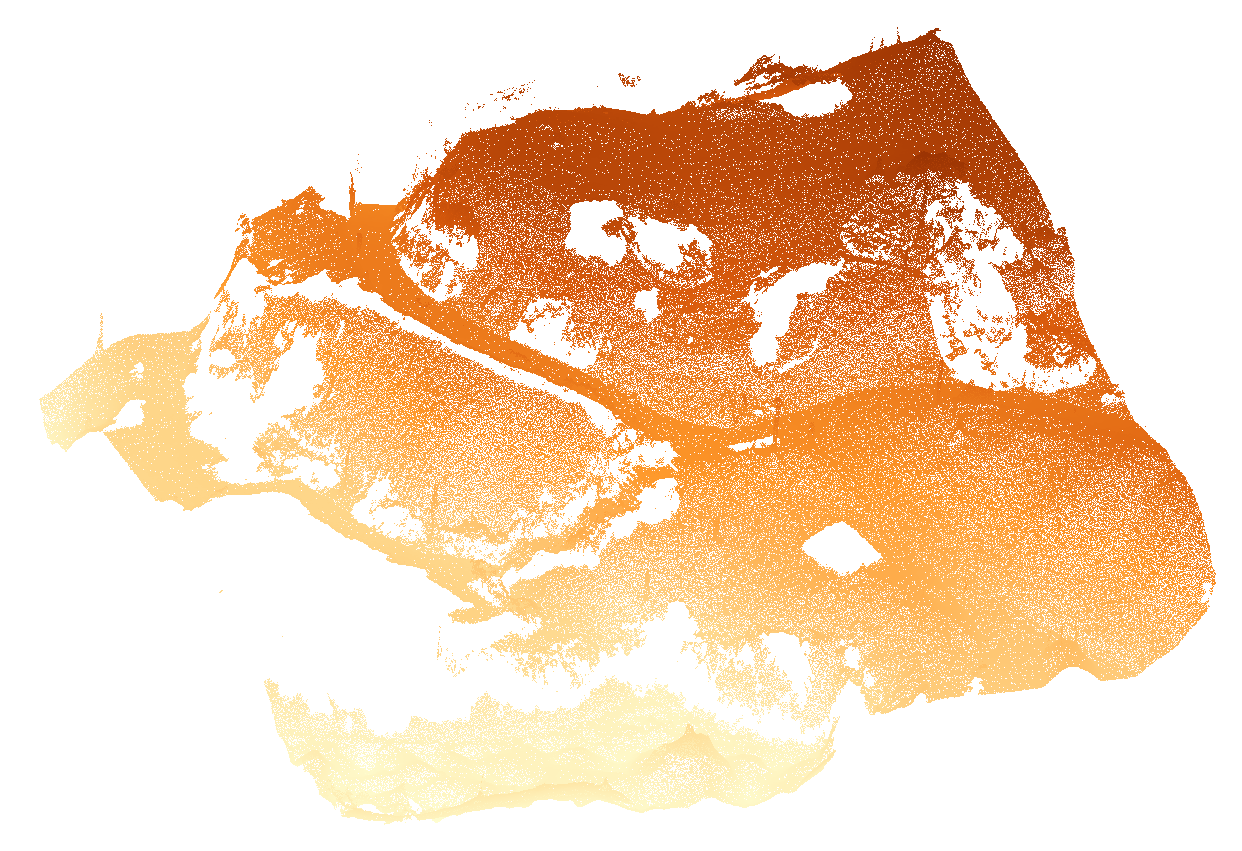}}
    \subfigure[DGI]{
    \includegraphics[width=0.47\linewidth]{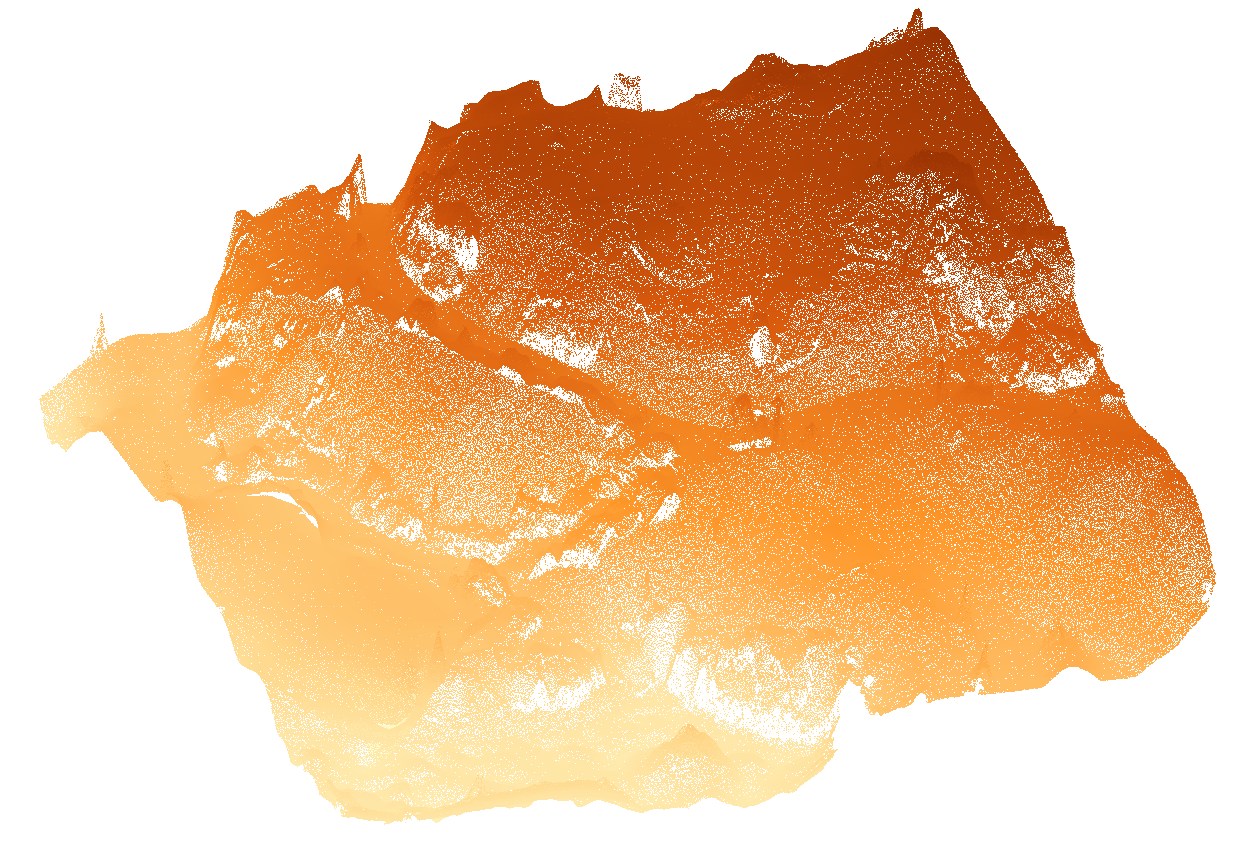}}
    \subfigure[MeshFix]{
    \includegraphics[width=0.47\linewidth]{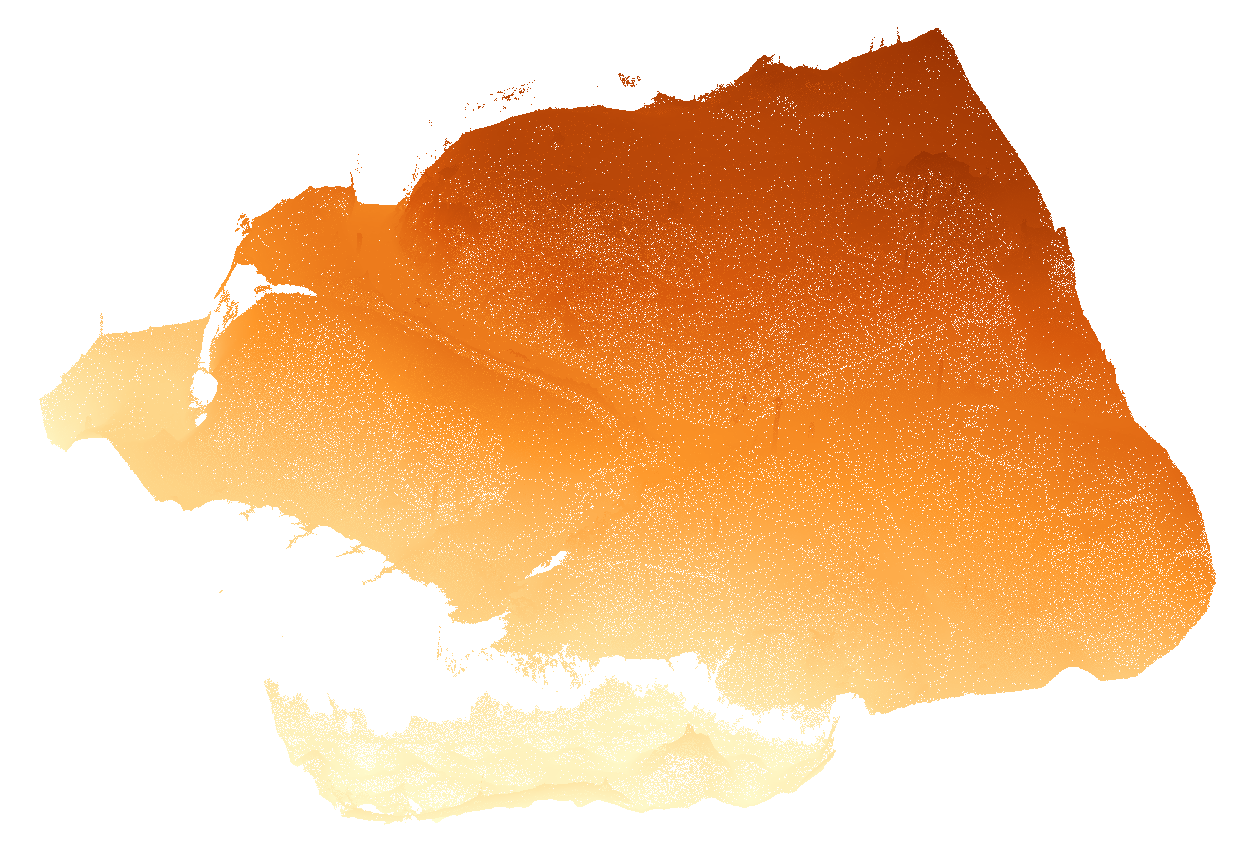}}
    \subfigure[Ours]{
    \includegraphics[width=0.47\linewidth]{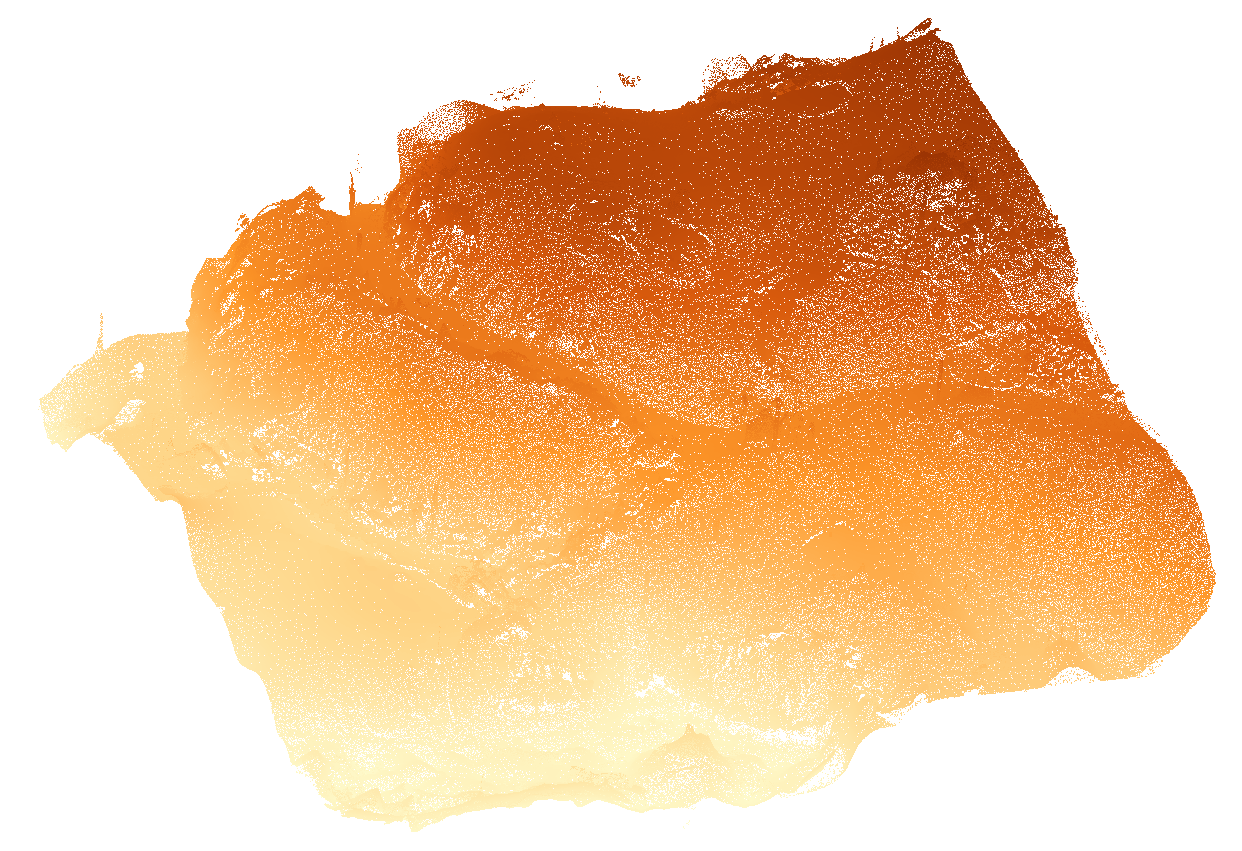}}
    \caption{Tests conducted on a real-world model without ground truth.}
    \label{Fig::real-world data}
\end{figure}

\begin{figure} [h]
    \includegraphics[width=1\linewidth]{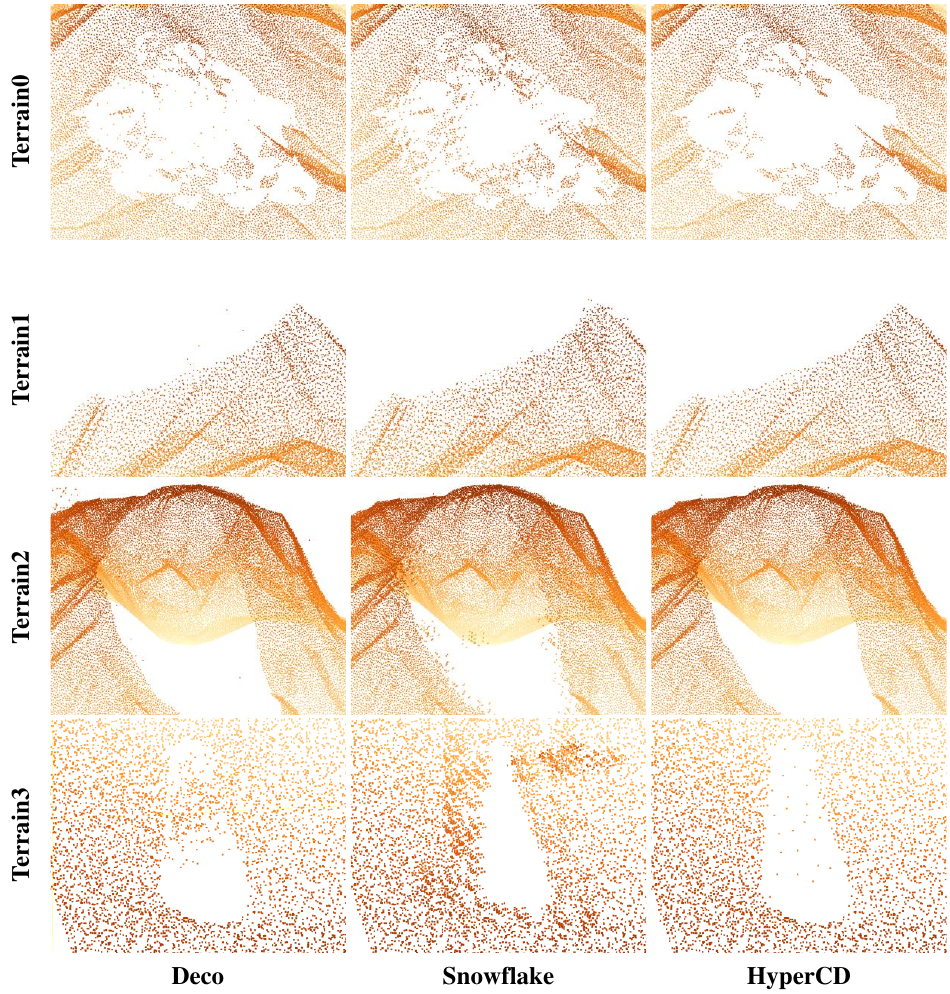}
\caption{We attempted to perform the inpainting task in our scenarios using deep learning-based algorithms. It is important to note that Deco produces partial inpainted point clouds as its output, whereas Snowflake and HyperCD generate complete inpainted point clouds. To facilitate a convenient comparison, we extracted points near the holes from their respective results at the same scale and merged them with input point clouds.}
\label{Fig::DL}
\end{figure}

From an objective perspective, we compare the results inpainted by ours and others with the ground truth and calculate two metrics: GPSNR and NSHD, respectively. As demonstrated in Table~\ref{Tab::GPSNR} and Table~\ref{Tab::NSHD}, our numerical values outperform those of other algorithms. Specifically, in terms of GPSNR, our method achieves an average score of 40.367DB, which is much higher than the others. Similarly, our solutions are also better in terms of NSHD, as we achieve lower values compared to the other methods. \par

Additionally, we explore deep learning-based methods for terrain point cloud inpainting. Due to limitations in the size of neural network inputs, we need to downsample our point cloud data. Subsequently, we split terrain point clouds into cubes and remove certain neighboring cubes to automatically generate a partial dataset for training purposes. The results depicted in Figure~\ref{Fig::DL} indicate that these methods cannot be directly applied to inpaint terrain scenes. This limitation arises from two key factors: (1) these methods are typically designed for smaller-scale point cloud data and demonstrate inadequate performance or in-applicability when applied to larger-scale point cloud data, and (2) these methods are commonly tailored to address substantial missing data, typically accounting for 50$\%$ or more of the point cloud, whereas the missing parts in our scenes are relatively small. The challenge lies in accurately localizing these holes, which ultimately hampers the effectiveness of these methods in the hole-filling process.

Moreover, we conduct a test in a scenario without ground truth, where holes are generated by employing a point cloud segmentation algorithm~\cite{khaloo2017robust} to extract and remove non-ground points. CSF and NBFM require the extraction of boundary points for the holes, which proves challenging in this specific scenario, resulting in their failure. Specifically, the former is unable to construct a complete boundary point loop that is essential for driving the inpainting process, while the latter, when substituting point cloud patches within the hole region, introduces additional boundary points as a consequence of imperfect registration, thus hindering the convergence of the inpainting process. As depicted in Figure~\ref{Fig::real-world data}, DGI exhibits significant information loss in the height dimension, leading to the presence of numerous hole regions in the vertical direction. Additionally, the large gaps between these hole regions result in incomplete mesh reconstruction when using MeshFix. In summary, our method surpasses others in terms of repair effectiveness and completion rate.

\section{Discussion}
\label{sec::discussion}

\begin{figure} [H]
    \includegraphics[width=1\linewidth]{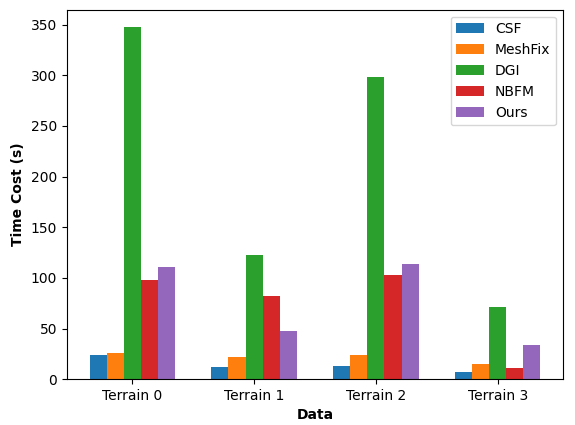}
\caption{Comparison with other algorithms in terms of time cost.}
\label{Fig::time cost}
\end{figure}

Certainly, our algorithm has its limitations. As illustrated in Figure~\ref{Fig::time cost}, it demonstrates disadvantages in terms of time complexity compared to alternative algorithms. This can be attributed to the significant time required for both B-spline fitting and solving the Poisson equation. The time complexity of fitting a uniform-parameterized cubic B-spline surface depends on factors such as the size of the point cloud data, the number of control points and iterations. Downsampling the point cloud provides an effective approach to accelerate the fitting process. The time complexity of solving the Poisson equation is influenced by the resolution of the height map, which is typically a high value to capture sufficient high-frequency information in large-scale scenarios. Additionally, solving large sparse linear systems using LU factorization has a time complexity of $O(n^3)$.

\section{Conclusion}
\label{sec::conclusion}

To address the presence of complex and indistinct boundary holes in terrain point clouds acquired through 3D acquisition technology, we propose a novel representation method for terrain point clouds to facilitate their inpainting. The fundamental idea is to decompose the point cloud signal into a low-frequency component represented by a B-spline surface and a high-frequency component represented by a relative height map generated from projecting discrete points to the surface. This transformation enables us to convert the point cloud inpainting problem into a surface fitting and 2D image inpainting problem. Subsequently, we combine the repaired high-frequency component with the complete low-frequency component to reconstruct new point clouds in 3D space. Additionally, we automate hole location based on the relative height map, eliminating the need for hole boundary extraction. \par

We conducted tests on several real-world scenes to evaluate our method, comparing it with some classical existing methods. The results demonstrate that our approach is capable of reconstructing more detailed geometric information and effectively handling complex hole boundaries. \par

Certainly, We have identified several areas for future work that warrant further investigation. Reducing time cost is part of our future research agenda, as we aim to address and overcome it. Considering the importance of these benchmarks, we also plan to further explore the application of our method on public release datasets in order to enhance its generalization. \par






\bibliographystyle{cag-num-names}
\bibliography{refs}



\end{document}